\documentclass{aims}
\usepackage{amsmath}
  \usepackage{paralist}
  \usepackage{graphics} %% add this and next lines if pictures should be in esp format
  \usepackage{epsfig} %For pictures: screened artwork should be set up with an 85 or 100 line screen
\usepackage{graphicx}  \usepackage{epstopdf}%This is to transfer .eps figure to .pdf figure; please compile your paper using PDFLeTex or PDFTeXify.
 \usepackage[colorlinks=true]{hyperref}
\hypersetup{urlcolor=blue, citecolor=red}

\def\currentvolume{X}
 \def\currentissue{X}
  \def\currentyear{200X}
   \def\currentmonth{XX}
    \def\ppages{X--XX}
     \def\DOI{10.3934/xx.xx.xx.xx}

\newtheorem{theorem}{Theorem}[section]

\theoremstyle{definition}

\newtheorem{remark}{Remark}

\newtheorem{assumption}{Assumption}

\newcommand{\eps}[1]{{#1}_{\varepsilon}}

\usepackage[usenames]{color}
\usepackage[final,notref,notcite]{showkeys}
\usepackage{subfigure}
\usepackage{multirow}
\usepackage[ruled,vlined]{algorithm2e}

\newcommand{\bm}[1]{\boldsymbol{#1}}

\newcommand{\st}{\text{ subject to }}

\newcommand{\mbfa}{\mathbf{A}}

\newcommand{\mbfd}{\mathbf{D}}

\newcommand{\mbfi}{\mathbf{I}}
\newcommand{\mbfm}{\mathbf{M}}

\newcommand{\mbfx}{\mathbf{X}}
\newcommand{\mbfy}{\mathbf{Y}}

\newcommand{\mbr}{\mathbb{R}}

\newcommand{\mca}{\mathcal{A}}

\newcommand{\mcc}{\mathcal{C}}
\newcommand{\mcd}{\mathcal{D}}

\newcommand{\mci}{\mathcal{I}}

\newcommand{\mcn}{\mathcal{N}}
\newcommand{\mcp}{\mathcal{P}}

\newcommand{\mcr}{\mathcal{R}}
\newcommand{\mcs}{\mathcal{S}}
\newcommand{\mct}{\mathcal{T}}

\newcount\refnum\refnum=0
\def\myref{{\global\advance\refnum by 1} {\bf \large Lecture \the \refnum. }}
\newcommand{\bfx}{{\bf x}}

\newcommand{\bfb}{{\bf b}}

\newcommand{\bfd}{{\bf d}}
\newcommand{\bfe}{{\bf e}}

\newcommand{\bfy}{{\bf y}}
\newcommand{\bfw}{{\bf w}}

\newcommand{\sign}{{\mathrm{sign}}}

\DeclareMathOperator*{\argmin}{argmin}

\usepackage[normalem]{ulem}

\title[Patch-dictionary method for whole image recovery]
      {A fast patch-dictionary method for whole image recovery}

\author[Y. Xu and W. Yin]{}


 \email{yangyang.xu@rice.edu}
 \email{wotaoyin@math.ucla.edu}

\thanks{Y. Xu is supported by NSF grant ECCS-1028790. W. Yin is partially supported by NSF grants DMS-0748839 and DMS-1317602, and ARO/ARL MURI grant FA9550-10-1-0567.}

\begin{document}

\maketitle

% Enter the first author's name and address:
\centerline{\scshape Yangyang Xu}
\medskip
{\footnotesize
% please put the address of the first author
 \centerline{ Department of Computational and Applied Mathematics, Rice University}
   \centerline{ Houston, TX 77005, USA}
} % Do not forget to end the {\footnotesize by the sign }

\medskip

\centerline{\scshape Wotao Yin}
\medskip
{\footnotesize
 % please put the address of the second  and third author
 \centerline{Department of Mathematics, University of California, Los Angeles, CA 90095, USA}
   %\centerline{}
}

\bigskip

% The name of the associate editor will be entered by an editorial staff
% "Communicated by the associate editor name" is not needed for special issue.
% \centerline{(Communicated by the associate editor name)}

%The abstract of your paper

\begin{abstract} Various algorithms have been proposed for dictionary learning. Among those for image processing, many use \emph{image patches} to form dictionaries. This paper focuses on whole-image recovery from corrupted linear measurements. %This includes tasks such as compressive sensing and medical image recovery.
We address the open issue of representing an image by \emph{overlapping} patches: the overlapping leads to an excessive number of dictionary coefficients to determine. % if they are computed the same time. 
With  very few exceptions, this issue has limited the applications of image-patch methods to the ``local'' kind of tasks such as denoising, inpainting, cartoon-texture decomposition, super-resolution, and image deblurring, for which one can process a few patches at a time. Our focus is global imaging tasks such as compressive sensing and medical image recovery, where the whole image is encoded together, making it either impossible or very ineffective to update a few patches at a time. 

Our strategy is to divide the sparse recovery into multiple subproblems, each of which handles a subset of non-overlapping patches, and then the results of the subproblems are averaged to yield the final recovery. This simple strategy is surprisingly effective in terms of both quality and speed.

%Most existing methods learn a dictionary from image patches, and thus the size of dictionary atoms is the same as that of image patches. The pioneering work \cite{elad2006image} uses a patch-size dictionary for image denoising. It motivates us to propose a non-overlapping patch-based method to recover an image from its linear measurements. To further improve recovery performance, we adaptively update the used dictionary after each image recovery process.

In addition, we accelerate  computation of the learned dictionary by applying a recent block proximal-gradient method, which not only has a lower per-iteration complexity but also takes fewer iterations to converge, compared to the current state-of-the-art. We also establish that our algorithm globally converges to a stationary point. Numerical results on synthetic data demonstrate that our algorithm can recover a more faithful dictionary than two state-of-the-art methods.

Combining our whole-image recovery and dictionary-learning methods,  we numerically simulate image inpainting, compressive sensing  recovery, and deblurring. % demonstrated that the learned dictionary by our algorithm works better than an overcomplete discrete cosine tranform, and our model works much better than
Our recovery is more faithful than those of a total variation method and a method based on overlapping patches. Our matlab code is competitive in terms of both speed and quality.
\end{abstract}

\section{Introduction}
Our general problem is to restore an image $\mbfm$ from its corrupted linear measurements in the form of $\bfb=\mca(\mbfm)+\bm{\xi}$, where $\mca$ is a linear operator and $\bm{\xi}$ is some noise. Examples of such recovery include image denoising ($\mca$ equals the identity operator $\mci$), super-resolution ($\mca$ is a downsampling operator), image deblurring ($\mca$ is a blurring operator),  compressive imaging recovery ($\mca$ is a compressed sensing operator), as well as medical imaging recovery ($\mca$ can be a downsampled Fourier or a Radon operator, for example).

This paper restores the image $\mbfm$ by computing its sparse representation under a learned dictionary. Following the approach pioneered in \cite{elad2006image}, we numerically form a dictionary that sparsely represents each and all the overlapping \emph{patches} of $\mbfm$. Given such a dictionary $\mbfd$, we  reconstruct the image patches  by finding their sparse coefficients and then recover the image from the patches.

We address an open issue regarding   \emph{whole--image} recovery: the large number of overlapping patches lead to a large number of free coefficients in the recovery, which can cause overfitting and  slow computation. This issue has  limited most of the patch-based methods (with a few exceptions we shall review below) to the ``local'' or ``nearly local'' kinds of image processing tasks such as denoising, inpainting, deblurring, and super-resolution. For these tasks, one or a few patches can be processed at a time, independently of the majority of the remaining patches, thus avoiding the overfitting issue. We, however, consider the more difficult ``global'' kind of task such as compressive sensing recovery, where each piece of the measurements encodes the whole image and thus it is either impossible or very ineffective to process one or a few patches at a time.

Bearing this issue in mind, we \emph{do not} process  either one patch at a time {or} all the overlapping patches at once, but instead   we process one subset of \emph{non-overlapping, covering} patches  at a time. (Covering means that the subset of patches covers all the pixels of the image.) Each time, we process this subset of patches and obtain a recovery of the whole image. After we process multiple different subsets of  \emph{non-overlapping, covering} patches, we obtain multiple whole-image recoveries, whose average is taken to eliminate the grid artifact that might exist in the individual ones. This simple strategy is surprisingly effective. Computationally, the different subsets of patches can be processed in parallel, and we found using merely five different subsets is enough to remove the grid artifact. For each subset, the corresponding  $\ell_1$ minimization problem is rather small: if $8\times 8$ patches are used, it only has roughly 1/64 of the free variables that one would have if all the {overlapping} patches are processed at once. Qualitatively, the averaged recovery has a higher PSNR than other state-of-the-art approaches that address the overfitting issue by applying either online optimization or incorporating additional image structures.

We also introduce a fast algorithm for learning the dictionary $\mbfd$, which plays a vital role in both our proposed recovery method and others. Here, $\mbfd$ can be pre-learned from a set of similar images, and then either fixed during the recovery or iteratively updated in adaptive to the image under recovery. Following \cite{elad2006image}, after recovering an image, we update the dictionary to fit the recovered image by solving an $\ell_1$-regularized model.
%the following model:
%$$
%\min_{\mbfd,\mbfy}~\mathcal{L}(\mbfd\mbfy,\mbfx)+\mathcal{R}(\mbfy)~ \st \|\bfd_i\|_2\le 1, i = 1,\ldots,K,
%$$
%where $\mathcal{L}$ is a differentiable data--fidelity function, $\mathcal{R}$ is an $\ell_1$-like sparsity promoting function, and $\bfd_i$ is the $i$th atom of dictionary $\mbfd$.
We introduce an algorithm to update dictionary $\mbfd$ and sparse coefficient $\mbfy$ alternatively. Unlike existing algorithms (e.g. \cite{engan2000multi, aharon2006rm}),  it does not exactly minimize over either $\mbfd$ or $\mbfy$, yet it decreases the energy very fast and provably converges to a stationary solution. Our code and several demos can be downloaded from our websites. Before giving more details of our approach and its numerical results, we first review the related literature.

\subsection{Image recovery by dictionary}
Various methods have been developed to restore an image from its corrupted and/or incomplete measurements.
%In general, they can be categorized into two classes. One class of methods perform recovery in the image space, i.e., treating image pixels as variables of the corresponding problem. Examples have the total variation (TV) based recovery \cite{Rudin-Osher-Fatemi-92, OsherBurgerGoldfarbXuYin2005} and spatial domain interpolation method for super-resolution \cite{keys1981cubic, park2003super}. Another class of methods do the recovery in a transformed space, and one most popular example is through sparse coding, which searches for sparse coefficients of the target image under a given sparsifying matrix or dictionary.
One popular class of recovery methods are based on sparse coding and dictionary such as those in \cite{elad2006image, anbarjafari2010image, ravishankar2011mr}.
% Once the sparse coefficients are found, the image itself can be easily restored by representing it as a combination of the given dictionary atoms using the recovered coefficients. Both kinds of methods have been shown efficient for natural image recovery. However, lots of works have demonstrated that given a ``nice'' dictionary, the second class of methods can outperform the first one in many cases; see \cite{elad2006image, anbarjafari2010image, ravishankar2011mr} for example. \sout{For this reason, we shall focus on the second approach and make a new algorithm to find a ``nice'' dictionary.} \commwy{In fact, there exist other image restoration methods, so this paragraph is not complete. So, I'd suggest removing it or reducing it to just 1 or 2 sentences and merge with the paragraph below.}
We say a signal $\bfx\in\mbr^n$ is sparse (or approximately sparse) under a dictionary $\mbfd\in\mbr^{n\times K}$ if $\bfx=\mbfd\bfy$ (or $\bfx\approx\mbfd\bfy$) and $\bfy\in\mbr^K$ has only a few nonzeros. Many types of signals can be sparsely represented by some dictionary. For example, natural images are approximately sparse under dictionaries based on various wavelet, curvelet, shearlet, and other transforms. Suppose $\bfx$ has a sparse representation under a dictionary $\mbfd$. Then given $\mbfd$ and linear measurements $\bfb=\mca(\bfx)+\bm{\xi}$, one can recover $\bfx$ through sparsely coding $\bfx$ via solving
\begin{equation}\label{spc}
\min_\bfy\|\bfy\|_0,\st \|\mca(\mbfd\bfy)-\bfb\|_2^2\le \eps,
\end{equation}
where $\|\cdot\|_0$ counts the nonzero number of its argument and is often approximated by $\|\cdot\|_1$ for tractable computation, and $\eps\ge 0$ is a parameter corresponding to $\bm{\xi}$. Once a solution $\bfy$ of \eqref{spc} is obtained, the original signal $\bfx$ can be estimated by $\mbfd\bfy$. The dictionary $\mbfd$ can be either predetermined or learned from a set of training data. Predetermined dictionaries, such as orthogonal or overcomplete wavelets, curvelets, and discrete cosine transforms (DCT), can have advantages of fast implementation over a learned one.
%\sout{ since the latter requires a large training dataset and often needs to solve a rather challenging optimization problem}.
Assuming easy availability of training datasets, however, it has been demonstrated (e.g., in \cite{kreutz2003dictionary, elad2006image, ramirez2010classification}) that a learned dictionary can better adapt to natural signals and improve the recovery quality.

For natural images, existing methods such as MOD \cite{engan2000multi} and KSVD \cite{aharon2006rm} learn a dictionary $\mbfd$ to sparsely represent the patches of an image, rather than the whole image itself. In other words, the size of dictionary atoms is the same as that of the image patches, for example, $6\times 6$ or $8\times 8$.
%\commwy{``Hence, $\mbfd\bfy$ can only represent an image patch but not the full image. Using models like \eqref{spc} to recover an image $\mbfm$ by a patch-size dictionary $\mbfd$, one has to apply some other representation of $\mbfm$ by $\mbfd$ or find another dictionary that is consistent with the full image.'' $\gets$ these sentences are not clear or accurate. I suggest removing them.}
To denoise an image $\mbfm$ with a patch-size dictionary, the pioneering work \cite{elad2006image} denoises each of the overlapping patches of $\mbfm$ via sparse coding and then estimates $\mbfm$ as the average of all the denoised patches together with the observed noisy image. This patch-based method was then extended to compressed sensing MRI -- a whole-image recovery problem -- in \cite{ravishankar2011mr}, which starts from a rough estimate of $\mbfm$, then simultaneously updates dictionary $\mbfd$ and sparse coefficients of all overlapping patches, and finally averages all the recovered patches
 to estimate $\mbfm$. 
%Liu et al. in \cite{liu2013highly} employ a similar model as that in \cite{ravishankar2011mr} but apply a different algorithm, called two-level Bregman method, to solve it. The work \cite{huang2013mr} further adds a total variation regularization term to improve the recovery quality. 
Dong et al. in \cite{dong2011image}
use local dictionaries to sparsely represent local patches and incorporate additional local auto-regression (AR) and non-local similarity (NLS) terms to reduce overfitting and improve recovery results.
%During the iterative procedure of solving the generalized model. which local dictionary is used for the $(i,j)$-th patch can be adaptively chosen, so can the AR and NLS terms.
Their model was demonstrated effective on image debluring and super-resolution. These and their follow-up works (e.g., \cite{fang2012sparsity, zhao2011hyperspectral}) use overlapping patches since tiling non-overlapping patches
%\sout{is observed to cause}
can cause visible grid artifact along the patch boundaries, which is avoided by using overlapping patches. %\sout{can remove these artifacts}.

\subsection{Learn a dictionary}
%\sout{While learned dictionaries have been shown more adaptive to natural signals and thus improve recovery performance of models like \eqref{spc}, efficiently solving the corresponding learning models becomes computationally challenging.}
Due to a lack of analytic structures, it can be computationally demanding to learn a dictionary. One of the most popular algorithms for dictionary learning is KSVD \cite{aharon2006rm}:
\begin{equation}\label{ksvd}
\min_{\mbfd,\mbfy}\|\mbfd\mbfy-\mbfx\|_F^2, \st \|\bfd_i\|_2=1, i = 1,\ldots,K;~ \|\bfy_j\|_0\le s, j=1,\ldots,p,
\end{equation}
where $\mbfx\in\mbr^{n\times p}$ is the training dataset, $\|\cdot\|_2$ denotes the Euclidean norm, $s$ is a parameter to control sparsity, and $\bfd_i$ is the $i$th column of $\mbfd$. KSVD attempts to solve \eqref{ksvd} by alternatively updating $\mbfy$ and $\mbfd$ in a certain way.
%\sout{Specifically, at each iteration, given $\mbfd$, it first performs sparse coding to each column $\bfx_j$ of $\mbfx$ by minimizing $\|\mbfd\bfy_j-\bfx_j\|_2$ with respect to $\bfy_j$ under constraint $\|\bfy_j\|_0\le s$. Then, it sequentially updates all columns of $\mbfd$, one at a time, simultaneously with the nonzero set of $\mbfy$ through rank-one matrix singular value decomposition (SVD). This method guarantees non-increasing monotonicity of the objective, and the learned dictionary has been demonstrated efficient for image denoising and inpainting. However, it is still unknown if the iterate sequence converges even to a stationary point.}
The objective is monotonically non-increasing and the denoising and inpainting performances are very good, but the convergence to a stationary point is not guaranteed. Furthermore, it is  slow as it performs SVD to update  $\mbfd$ and exact minimization to update every $\bfy_j$ in each iteration.
%\sout{performing sparse coding and SVD at each iteration makes the method computationally very expensive, in particular when the training data has great many, say millions of, samples.}

Another popular method is the online dictionary learning (OLM) \cite{mairal2009online}, which, via an online update approach, attempts to solve
\begin{equation}\label{spdict}
\min_{\mbfd,\mbfy}\frac{1}{2}\|\mbfd\mbfy-\mbfx\|_F^2+\lambda\|\mbfy\|_1, \st \|\bfd_i\|_2\le 1, i = 1,\ldots,K,
\end{equation}
where $\|\mbfy\|_1=\sum_{i,j}|y_{ij}|$ is a convex relaxation of $\|\cdot\|_0$, and $\lambda$ is a tuning parameter to balance data fitting and sparsity level. OLM alternatively updates $\mbfy$ and $\mbfd$ as follows.
%\sout{as KSVD does, but uses different updates. Specifically,}
When $\mbfd$ is fixed, it randomly picks a batch of columns of $\mbfx$ and applies sparse coding to each selected column. %$\bfx_j$. %by minimizing $\frac{1}{2}\|\mbfd\bfy_j-\bfx_j\|_2^2+\lambda\|\bfy_j\|_1$ with respect to $\bfy_j$.
Letting $S$ be the index set of all previously selected samples and $\mbfy_S$ contain their sparse coefficients, the method then updates $\mbfd$ to the solution of $\min_{\mbfd}\{\|\mbfd\mbfy_S-\mbfx_S\|_F^2,\|\bfd_i\|_2\le 1,\forall i\}$, where $\mbfx_S$ denotes the submatrix consisting of all columns of $\mbfx$ indexed by $S$. The above two steps are then repeated until convergence. The algorithm often runs faster than KSVD, and its efficiency relies on the assumption that all training samples have the same distribution. Assuming that the training data admits bounded probability with a compact support and $\mbfy_S\mbfy_S^\top$ is uniformly positive definite, it is shown that the iterate sequence asymptotically satisfies the first-order optimality condition of \eqref{spdict}. The global convergence of the iterate sequence is still open.

%\sout{Some other algorithms like MOD [9] %\cite{engan2000multi}
%can also produce similar results as KSVD or OLM.}
We refer the interested readers to the review paper \cite{tosic2011dictionary} for other dictionary learning methods. In addition, more complicated models have been proposed to learn dictionaries for specific tasks; see \cite{mairal2008supervised, mairal2012task} for example. We do not intend to consider those models and will keep our focus on \eqref{spdict} in this paper.

\subsection{Contributions}
This paper makes the following contributions: %in all of the three aspects: modeling, algorithm and application.
\begin{itemize}
\item We propose a simple, novel method that recovers a whole image by applying sparse coding to its patches. In addition to the traditional denoising, inpainting, and deblurring tasks, the method can be applied to recovering an image from its whole-image linear measurements, which arise in the applications of compressive sensing and medical imaging. The method is simple and can include additional energy terms and constraints, as well as to be embedded in more complicated imaging applications.
We want to emphasize that our method recovers the whole image at a time and is different from local recovery methods such as those in \cite{elad2006image, mairal2009online} which process image patches one by one.

\item Along with the method, we introduce a numerical algorithm for  dictionary learning that is fast and has provable convergence to a stationary point. The algorithm is based on our recent work on block proximal gradient update in \cite{xublock}. Compared to the existing algorithms, the proposed algorithm has a low per-iteration cost and converges fast. %\sout{In addition, the iterate sequence of this method is shown globally convergent to a stationary point. To the best of our knowledge, this global convergence property is not owned by other dictionary learning methods. Numerical results show that our algorithm is not only faster but also produces better solutions than some state-of-the-art methods.}

\item We provide Matlab codes for three different imaging tasks that are
%\sout{Our image recovery procedure enables whole-image compressive sensing and other tasks that recover an image as a whole to take advantage of patch-based image sparse coding. We test the proposed procedure and dictionary learning algorithm for three different image recovery tasks.}
(i) inpainting: fill in image missing pixels; (ii) compressive sensing recovery: recover an image from its undersampled linear measurements; (iii) image deblurring:  restore a clean image from its blurs. %\sout{All the three examples have many applications and can be easily realized in hardware.}
%\commwy{``easy hardware realization'' is interesting; any justifications?} %\sout{Numerical experiments show that our procedure using the dictionary learned by the proposed algorithm performs consistently better than both TV-based method and the overlapping patch-based method of \cite{dong2011image}.}
On these tasks, our codes compare favorably to total variation (TV) methods, as well as those from \cite{mairal2009online,dong2011image} using overlapping patches and learned dictionaries.
\end{itemize}

\subsection{Organization}
The rest of the paper is organized as follows. In section \ref{sec:rec}, we give a new model for recovering an image from its linear measurements, and also discuss how to improve recovery results. Section \ref{sec:alg} applies a block proximal gradient method to \eqref{spdict} and makes a new dictionary learning algorithm. Numerical results are reported in section \ref{sec:numerical}, and finally section \ref{sec:conclusion} concludes the paper.

\section{Problem formulation}\label{sec:rec}
Given a patch-size dictionary $\mbfd$, we aim at recovering an image $\mbfm$ from its corrupted linear measurements $\bfb=\mca(\mbfm)+\bm{\xi}$, where $\mca$ is a linear operator and $\bm{\xi}$ is some noise. The case of $\mca=\mci$ has been considered in the pioneering work \cite{elad2006image}, which alternatively performs sparse coding to denoise every patch and takes average over all overlapping denoised patches together with the observed noisy image.
% $\hat{\mbfm}=\mbfm+\bm{\xi}$ by
%\begin{equation}\label{average}\mbfm=\big(\mu\mci+\sum_{(i,j)\in S}\mcr_{ij}^\top\mcr_{ij}\big)^{-1}\big(\mu\hat{\mbfm}+\sum_{(i,j)\in S}\mcr_{ij}^\top(\mbfd\bfy_{ij})\big),
%\end{equation}
%where $\mu$ is a parameter corresponding to $\bm{\xi}$, $S$ denotes the set of all overlapping patches of $\mbfm$, $\bfy_{ij}$ is the sparse coefficient of the $(i,j)$-th image patch under $\mbfd$, $\mcr_{ij}$ is an operator taking the $(i,j)$-th patch, and $\mcr_{ij}^\top$ is the adjoint of $\mcr_{ij}$. Note that $\sum_{(i,j)\in S}\mcr_{ij}^\top\mcr_{ij}$ is diagonal, and thus the inverse operator in \eqref{average} can be implemented in a pixel-by-pixel manner.

Throughout the discussion in the remaining part of the paper, we assume that a generic image has size $N_1\times N_2$ and training patches to be $n_1\times n_2$. The dictionary $\mbfd$ has $K$ atoms, and all of them are vectors in $n_1n_2$ dimensional space. Keep in mind that an $m\times n$ matrix is equivalent to an $m\cdot n$ vector under Matlab's \verb|reshape| operation. Hence, we will use a matrix and its reshaped vector interchangeably. For example, a dictionary atom can be regarded as either a vector of length  $n_1 n_2$ or an $n_1\times n_2$ patch.

\subsection{Our model}
Motivated by \cite{elad2006image}, we exactly represent an image by $$\mbfm=\big(\mct_P\big)^{-1}\big(\sum_{(i,j)\in P}\mcr_{ij}^\top\mcr_{ij}(\mbfm)\big),\quad \mct_P:=\sum_{(i,j)\in P}\mcr_{ij}^\top\mcr_{ij}$$
where $\mcr_{ij}$ is an operator taking the $(i,j)$-th patch, $\mcr_{ij}^\top$ is the adjoint of $\mcr_{ij}$, and $P$ contains a subset of patches covering all the pixels of $\mbfm$, ensuring that $\mct_P$ is invertible. Note that $\mct_P$ is diagonal, and thus its inverse can be implemented in a pixel-by-pixel manner. If every patch $\mcr_{ij}(\mbfm)$ in $P$ has a sparse representation under $\mbfd$, i.e., $\mcr_{ij}(\mbfm)=\mbfd\bfy_{ij}$ for a sparse vector $\bfy_{ij}$, then the above representation can be written as
\begin{equation}\label{aver_im}\mbfm=\big(\mct_P\big)^{-1}\big(\sum_{(i,j)\in P}\mcr_{ij}^\top(\mbfd\bfy_{ij})\big).
\end{equation}
Using this representation, we make the following weighted $\ell_1$ model:
\begin{equation}\label{con-model}
\min_\bfy\sum_{(i,j)\in P}\|\bfw_{ij}\odot\bfy_{ij}\|_1,\st \big\|\mca\mct_P^{-1}\big(\sum_{(i,j)\in P}\mcr_{ij}^\top(\mbfd\bfy_{ij})\big)-\bfb\big\|_2\le \sigma,
\end{equation}
where $\bfw_{ij}\ge0$ is a weight vector for $(i,j)\in P$, $\sigma$ is the noise level determined by $\bm{\xi}$, and ``$\odot$'' denotes component-wise product. Equivalently, one can consider the unconstrained model:
\begin{equation}\label{un-model}
\min_\bfy\sum_{(i,j)\in P}\|\bfw_{ij}\odot\bfy_{ij}\|_1+\frac{1}{2\nu}\big\|\mca\mct_P^{-1}\big(\sum_{(i,j)\in P}\mcr_{ij}^\top(\mbfd\bfy_{ij})\big)-\bfb\big\|^2_2,
\end{equation}
where $\nu$ is a parameter corresponding to $\sigma$. Upon solving \eqref{con-model} or \eqref{un-model}, one can use $\mct_P^{-1}\sum_{(i,j)\in P}\mcr^\top_{ij}(\mbfd\bfy_{ij})$ to estimate $\mbfm$.

\begin{remark}
%Following \cite{elad2006image}, Dong et al. \cite{dong2011image} represent the image $\mbfm$ by
%\begin{equation}\label{repM}\mbfm=\big(\sum_{(i,j)\in S}\mcr_{ij}^\top\mcr_{ij}\big)^{-1}\big(\sum_{(i,j)\in S}\mcr_{ij}^\top\mcr_{ij}(\mbfm)\big),
%\end{equation}
%and they consider general $\mca$.
%Instead of using a global dictionary, they propose to use local dictionaries and further incorporate AR and NLS terms to make the model
Our models are similar to that in \cite{dong2011image}:
\begin{equation}\label{asds}
\begin{array}{l}\underset{\bfy}{\min} \underset{(i,j)\in S}{\sum}\|\bfy_{ij}\|_1+\frac{1}{2\nu}\left\|\mca\left(\big(\underset{(i,j)\in S}{\sum}\mcr_{ij}^\top\mcr_{ij}\big)^{-1}\big(\underset{(i,j)\in S}{\sum}\mcr_{ij}^\top(\mbfd_{k_{ij}}\bfy_{ij})\big)\right)-\bfb\right\|_2^2\\[0.3cm]
\hspace{1cm}+\text{AR}(\bfy)+\text{NLS}(\bfy),
\end{array}
\end{equation}
where $S$ denotes the set of all overlapping patches, $\nu$ is a parameter balancing sparsity and data fitting, $\mbfd_{k_{ij}}$ is a given local dictionary used to represent the $(i,j)$-th patch, and $\text{AR}(\cdot)$ and $\text{NLS}(\cdot)$ are two regularization terms corresponding to local auto-regression and non-local similarity.
%While solving \eqref{asds}, they adaptively choose a local dictionary to represent the $(i,j)$-th patch, i.e., the index $k_{ij}$ is updated. Note that a global dictionary $\mbfd$ as in \eqref{average} is often overcomplete (i.e., more columns than rows) while in \eqref{asds}
The local dictionaries are often incomplete (i.e., fewer columns than rows). Similar to non-overlapping patches (see next paragraph), non-completeness of local dictionaries and AR and NLS terms can reduce variable freedom and increase recoverability of \eqref{asds}. However, the use of more dictionaries and complicated regularization terms makes \eqref{asds} more difficult to solve than our models.
\end{remark}

\subsubsection*{Choice of $P$} 
\begin{figure}\caption{Image denosing comparison of two methods: (left image) solving \eqref{un-model} with all patches used at once, (right image) solving \eqref{un-model} with one subset of non-overlapping, covering patches.
%, (right image) applying \eqref{average} with $\mu=0$.
In \eqref{un-model},  $\nu=0.05$, and $\mbfd$ was learned according to section \ref{sec:imgrec}. %We employed YALL1 (version 1.4) \cite{zhang2010yall1} to solve \eqref{un-model}, and the parameters of YALL1 were set in the same way as in section \ref{sec:imgrec}.
}\label{fig:overlap}
\begin{center}
\begin{tabular}{cc}
\includegraphics[width=0.3\textwidth]{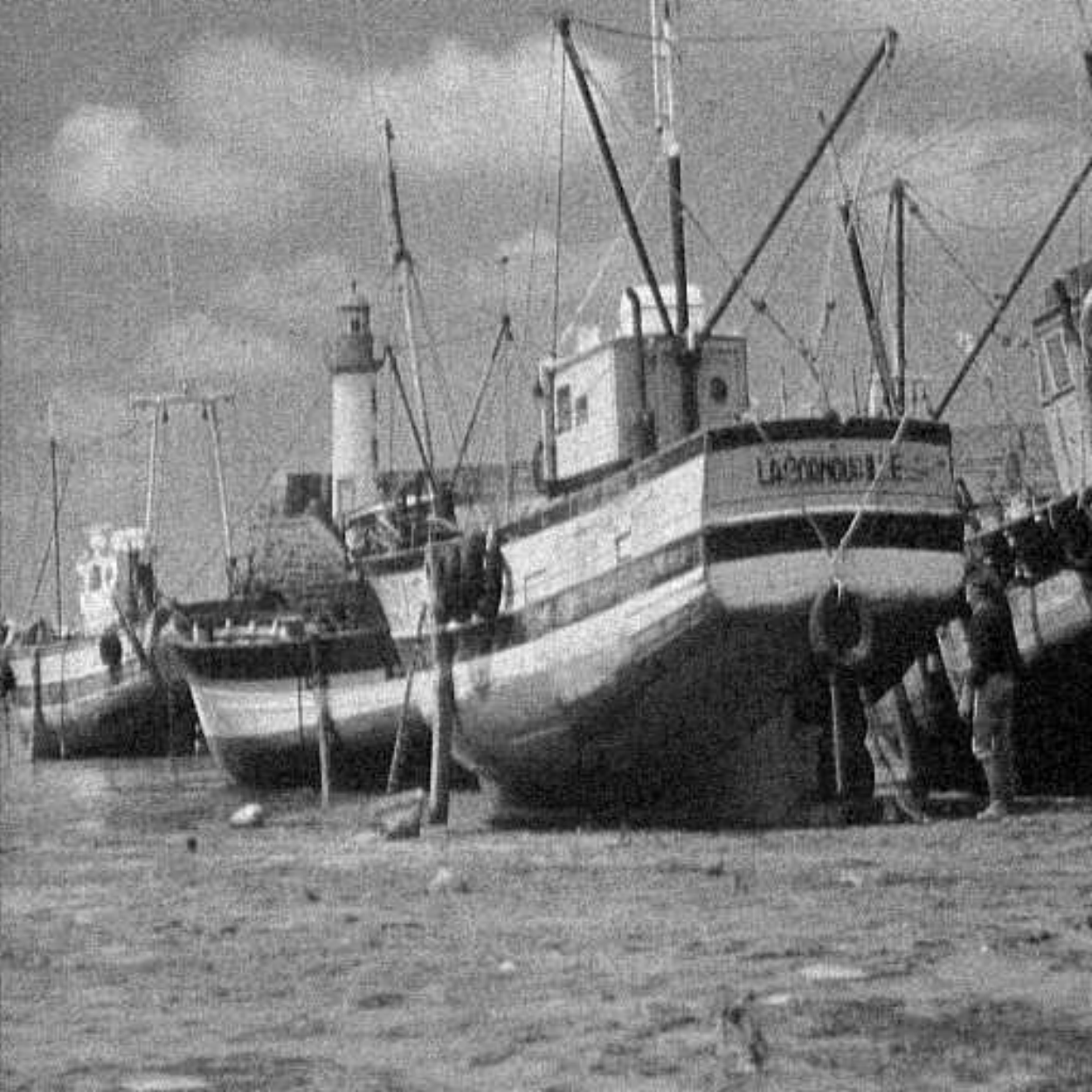} &
\includegraphics[width=0.3\textwidth]{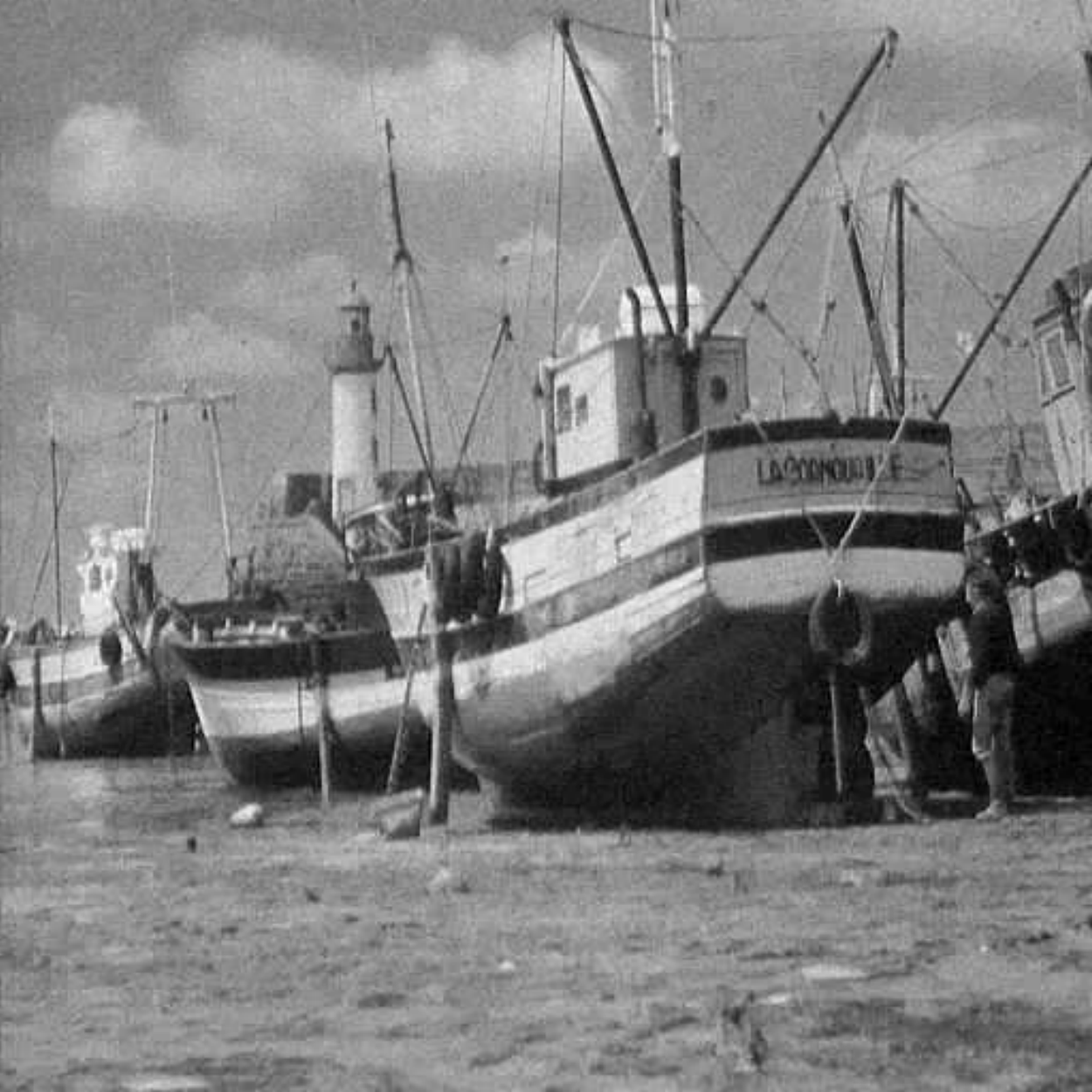} \\
PSNR = 26.98 & PSNR = 30.57\\
All patches used at once & One subset of non-overlapping\\
& covering patches
\end{tabular}
\end{center}
\end{figure}

One question is how to choose $P$, the subset of covering patches, such that \eqref{con-model} or \eqref{un-model} work well for recovering $\mbfm$. We let  $P$ be a subset of non-overlapping, covering patches and focus on the unconstrained model \eqref{un-model}. Figure \ref{fig:overlap} compares the two approaches. In this test, we set $\mca=\mci$ and $\bfb=\mbfm+0.05\bm{\xi}$ with $\bm{\xi}\sim\mcn(0,\mbfi)$, and we compared \eqref{un-model} with two different $P$'s. In Figure \ref{fig:overlap}, the left image uses all overlapping patches, and the right image uses one subset of non-overlapping, covering patches. We see that \eqref{un-model} with all patches produces much worse result than that with non-overlapping $P$.

We want to emphasize here that our results do not counter the intuition that using more patches should give better recovery. The results in Table \ref{table:avg} of section \ref{sec:numerical} demonstrate that using more different subsets of non-overlapping, covering patches can consistently improve the recovered image quality. The phenomenon in Figure \ref{fig:overlap} can be explained as follows.
Using all the overlapping patches in \eqref{con-model} or \eqref{un-model} introduces too many unknowns to decide.
%\sout{for which case the method in \cite{elad2006image} performed very well. Intuitively, using all patches would make better recovery. However, it makes excessively many unknowns.}
The $\ell_1$ minimization typically needs  $O(s\log (n/s))$ or more measurements to recover an $s$-sparse signal of length $n$. Suppose that the $\bfy_{ij}$ corresponding to each patch has at least $r$ nonzeros and all the $(N_1-n_1+1)(N_2-n_2+1)$ overlapping patches are used. Then vector $\bfy$ has $n=K(N_1-n_1+1)(N_2-n_2+1)$ entries out of which at least $s=r(N_1-n_1+1)(N_2-n_2+1)$ are nonzeros. On the other hand, we have at most $N_1N_2$ measurements, not sufficiently many to reach $O(s\log (n/s))=O(rN_1N_2\log(K/r))$. \emph{Therefore, unless more constraints or regularizations on $\bfy$ are introduced to help (see \eqref{asds} for instance), we cannot use all the patches.}
%as did in \eqref{asds}.
%We choose the former approach to address the trouble caused by too many unknowns.

\begin{figure}\caption{Three different ways of partitioning a $100\times 100$ image into non-overlapping patches, where each patch is no greater than $8\times 8$ and all interior patches have size $8\times 8$.}\label{fig:partition}
\begin{center}
\begin{tabular}{ccc}
partition 1 & partition 2 & partition 3\\
\includegraphics[width=0.31\textwidth]{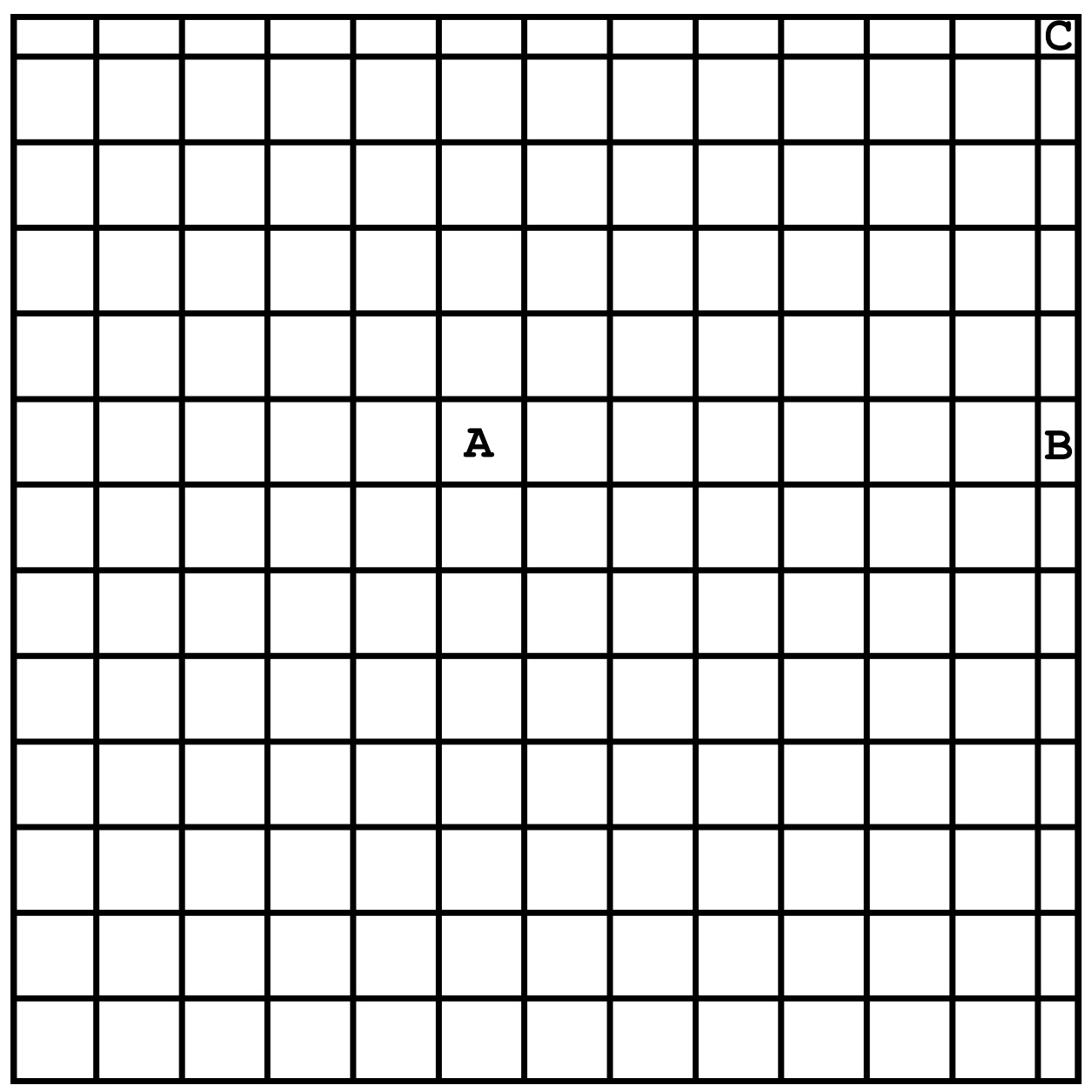} &
\includegraphics[width=0.31\textwidth]{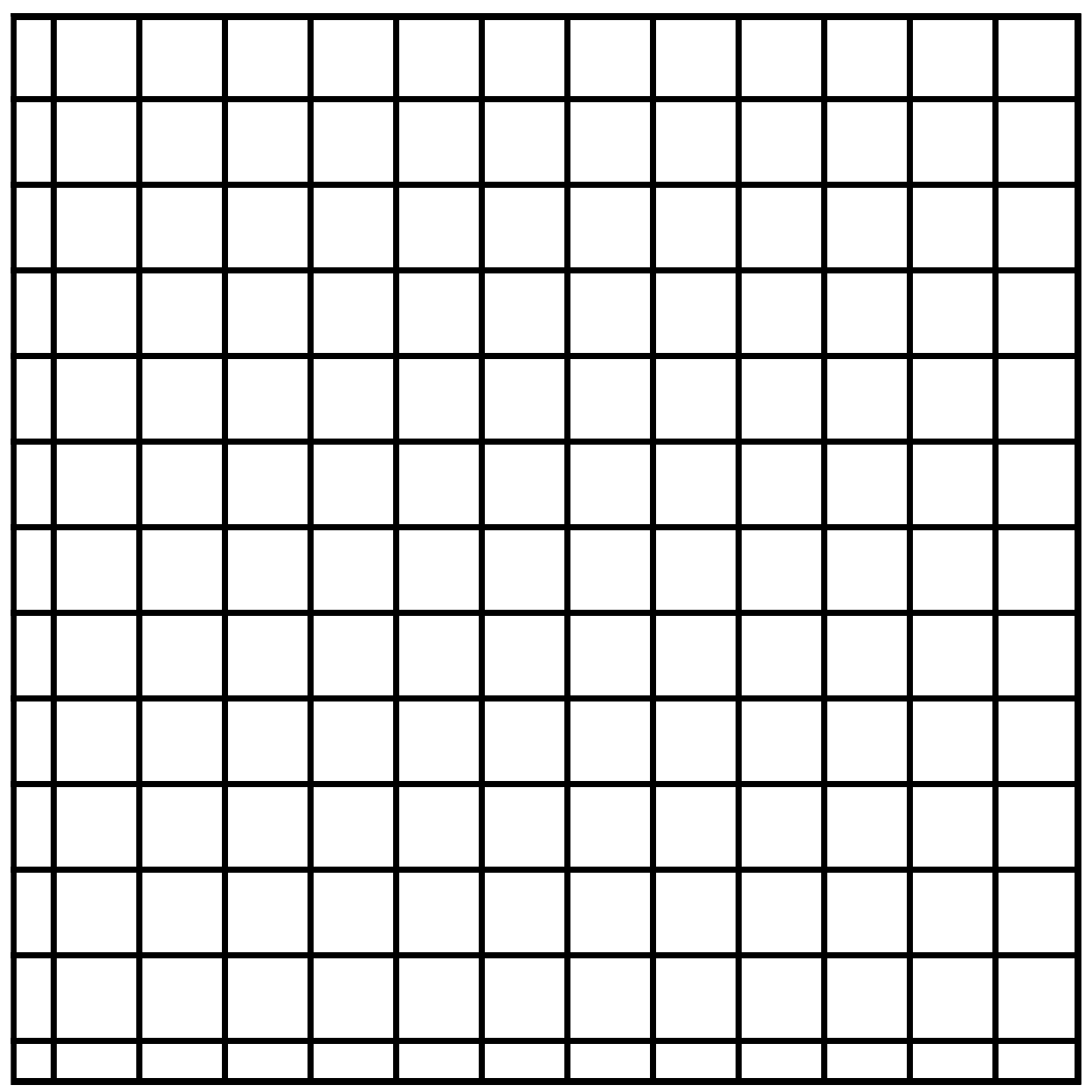} &
\includegraphics[width=0.31\textwidth]{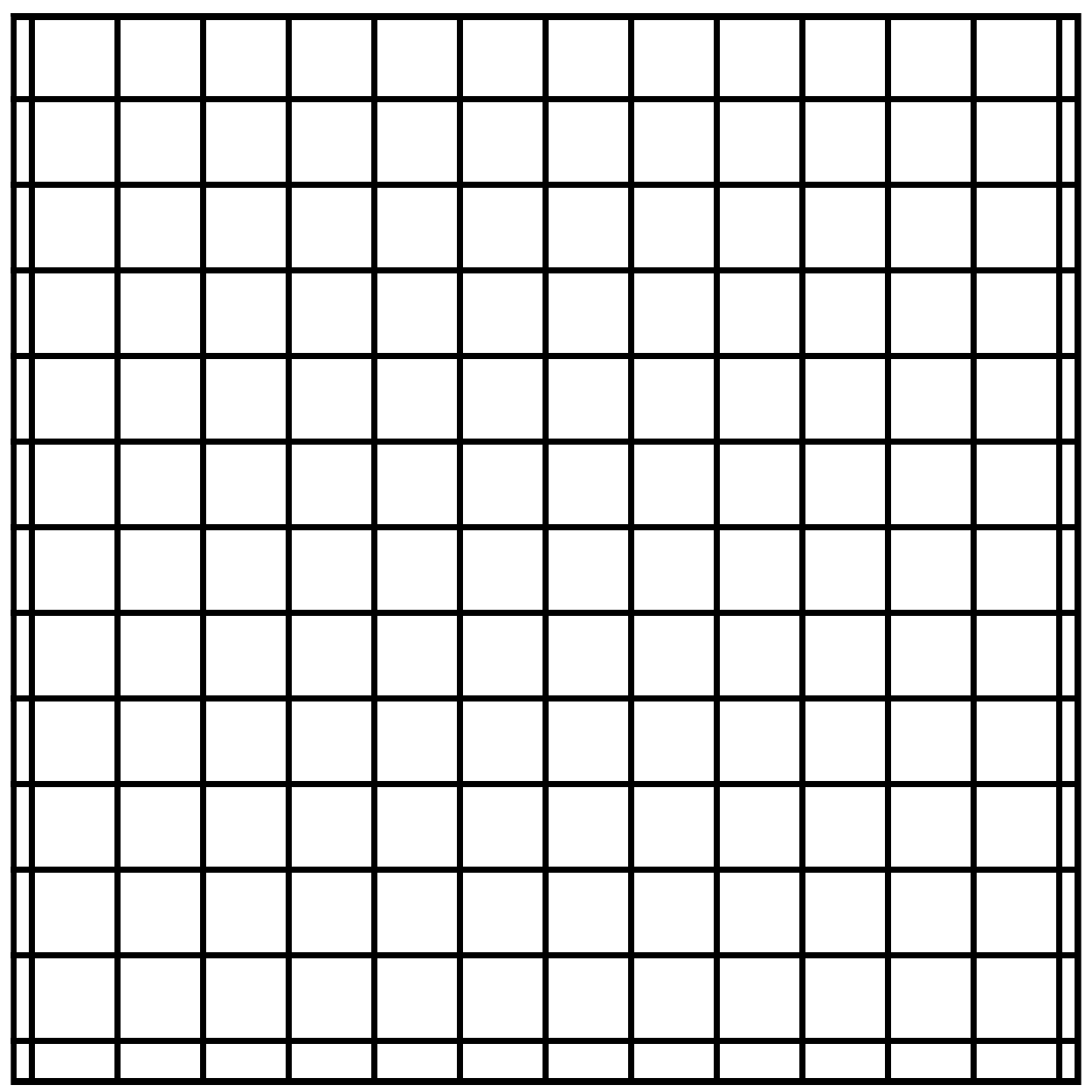}
\end{tabular}
\end{center}
\end{figure}

Since the image may not be evenly divided and the selected patches need to cover all the pixels of the image, we allow them to have different sizes. Slightly abusing the notation, we still use $P$ to denote the set of selected patches, but
%\sout{note that $P$ may not be a subset of $S$ since}
$P$ can also contain some smaller patches near the boundary. Although we can partition the image arbitrarily with blocks no greater than $n_1\times n_2$, for simplicity we make the following assumptions.
\begin{assumption}[Image partition by patches%Method of partitioning an image
]\label{assump:partition}\ \\[-0.4cm]
\begin{itemize}
\item Interior patches (e.g., patch  ``A'' in Figure \ref{fig:partition}) have size $n_1\times n_2$;
\item Left and right boundary patches have $n_1$ rows, and lower and upper boundary patches have $n_2$ columns; patch ``B'' in Figure \ref{fig:partition} is an example;
\item Corner patches (e.g., patch ``C'' in Figure \ref{fig:partition}) can have fewer than $n_1$ rows and $n_2$ columns;
\item All patches are vertically and horizontally aligned. % in the same row of a partition have the same number of rows, e.g., in the first row of partion 1 of Figure \ref{fig:partition}, all patches have 4 rows;
%\item All patches in the same column of a partition have the same number of columns; e.g., in the last column of partion 1 of Figure \ref{fig:partition}, all patches have 4 columns.
\end{itemize}
\end{assumption}

\begin{remark}
Under Assumption \ref{assump:partition}, the way an image is partitioned into patches is  uniquely determined by the size of the upper-left corner patch.
\end{remark}

 Figure \ref{fig:partition} illustrates how we partition a $100\times 100$ image into non-overlapping patches in three different ways. Every patch is no greater than $8\times 8$, and all interior patches are $8\times 8$. However, since the image cannot be evenly partitioned, the patches near the boundary of the image may be smaller than $8\times 8$. For example, in partition 1, all the right boundary patches are $8\times 4$, and the upper-right corner patch  is $4\times 4$; in partition 3, all the left and right boundary patches are $8\times 2$, and the lower-left and lower-right corner patches are $4\times 2$.
%In our numerical experiments, even when the image can be evenly partitioned, we will still choose some uneven way to partition it.

%Below we use a global dictionary to make a model, which is similar to \eqref{asds} but uses non-overlapping patches and does not contain AR and NLS terms. Restricting image patches to non-overlapping ones can significantly decrease the number of unknows to a level that compressed sensing theory \cite{candes2008introduction} guarantees robust recovery if the measurements are sufficiently many. In addition, our model becomes easier to solve than overlapping patch-based ones such as \eqref{asds}.

\subsubsection*{Definition of operators}
As $P$ consists of non-overlapping covering patches, then every pixel must be contained by exactly one patch, and it is not difficult to verify $\mct_P=\mci$. If $(i,j)\in P$ is one interior patch, then $\mcr_{ij}(\mbfm)$ means to take the $(i,j)$-th patch of $\mbfm$, and $\mcr_{ij}^\top(\bfx)$ is to first generate an $N_1\times N_2$ zero matrix, and then add $\bfx$ to its $(i,j)$-th patch. However, as $(i,j)\in P$ is a boundary or corner patch and its size is smaller than $n_1\times n_2$, the corresponding operators need to act accordingly.
%For example, let $(i,j)$ be the upper-left $4\times 8$ corner patch in partition 1 of Figure \ref{fig:partition}. Then we define
%\begin{itemize}
%\item $\mcr_{ij}(\mbfm)$: first generate an $8\times 8$ zero matrix, and then replace its \emph{first four} rows with the \emph{upper-left} $4\times 8$ corner patch of $\mbfm$;
%\item $\mcr_{ij}^\top(\bfx)$: first generate a $100\times 100$ zero matrix, and then replace its \emph{upper-left} $4\times 8$ corner patch with the \emph{first four} rows of $\bfx$.
%\end{itemize}
For example, let $(i,j)$ be patch ``C'' in Figure \ref{fig:partition}. Then we define
\begin{itemize}
\item $\mcr_{ij}(\mbfm)$: first generate an $8\times 8$ zero matrix, and then replace its \emph{upper-right} $4\times 4$ corner submatrix with the \emph{upper-right} $4\times 4$ corner patch of $\mbfm$;
\item $\mcr_{ij}^\top(\bfx)$: first generate a $100\times 100$ zero matrix,  and then replace the \emph{upper-right} $4\times 4$ corner patch corresponding to ``C'' with the \emph{upper-right} $4\times 4$ corner submatrix of $\bfx$.
\end{itemize}

\subsubsection*{Averaging scheme}
As shown in \cite{elad2006image}, tiling non-overlapping patches to perform image denoising would yield visible artifacts on block boundaries, and it was also observed when we solved \eqref{un-model} once with non-overlapping patches. Though using all patches in \eqref{un-model} at once does not give good recovery, we still want to use them in some way.  Note that we have the freedom to choose $P$ in \eqref{un-model}, so we can solve it for different $P$'s. For example, if $\mbfm\in\mbr^{100\times 100}$ and dictionary atoms are $8\times 8$, we can partition the image into non-overlapping patches in the three different ways in Figure \ref{fig:partition}, and solve \eqref{un-model} for each partition. It turns out that averaging the recovered images from different $P$'s can remove the artifacts occuring on block boundaries and improve PSNR value; see the numerical results in section \ref{sec:numerical}. Algorithm \ref{alg:rec} summarizes our method. Note that \eqref{un-model} can be solved for different $P$'s in parallel.

\begin{algorithm}\caption{}\label{alg:rec}
{\small
\DontPrintSemicolon
\KwData{Dictionary $\mbfd$, patch size $(n_1,n_2)$, image size $(N_1,N_2)$, measurements $\bfb$, linear operator $\mca$, and parameter $\nu$.}
{\bf Choose} $t$ different ways to partition the image into non-overlapping patches; denote them as $P_1,\ldots,P_t$.\;
{\bf Solve} \eqref{un-model} for $P_k$ and let the recovered image be $\mbfm_k$, for $k=1,\ldots,t$. \;
{\bf Average} all the recovered images by $\tilde{\mbfm}=\frac{1}{t}\sum_{k=1}^t\mbfm_k$ and output $\tilde{\mbfm}$.
}
\end{algorithm}

\subsection{Adaptive dictionary update}\label{sec:adp}
After obtaining an estimated image $\tilde{\mbfm}$ by Algorithm \ref{alg:rec}, we can update the dictionary $\mbfd$ using patches extracted from $\tilde{\mbfm}$. Since $\tilde{\mbfm}$ is close to the original image $\mbfm$, the updated dictionary $\mbfd$ from $\tilde{\mbfm}$ should better represent the patches of $\mbfm$. Hence, it is possible to further improve the result using the adaptively updated dictionary, and this process can be repeated several times. Algorithm \ref{alg:adp} summarizes our adaptive method.

We observe that only the first adaptive update gives significant improvement, and subsequent ones make only minor changes to the dictionary and thus little improvement to the recovered image. For this reason, in the numerical experiments, we will update the dictionary only once.

\begin{algorithm}\caption{}\label{alg:adp}
{\small
\DontPrintSemicolon
\KwData{Dictionary $\mbfd$, patch size $(n_1,n_2)$, image size $(N_1,N_2)$, measurements $\bfb$, linear operator $\mca$, and parameter $\nu$.}
\Repeat{convergence}{
{\bf Run} Algorithm \ref{alg:rec} and let the recovered image be $\tilde{\mbfm}$. \;
{\bf Update} dictionary $\mbfd$ from patches extracted from $\tilde{\mbfm}$.
}
}
\end{algorithm}

\section{Block proximal gradient method for dictionary learning}\label{sec:alg}
Both Algorithms \ref{alg:rec} and \ref{alg:adp} require an initial dictionary $\mbfd$, which can be an analytic dictionary such as orthogonal or overcomplete wavelets, curvelets or DCT, or a learned one. For our purpose, a learned dictionary is preferable since it can be more adaptive to natural images \cite{kreutz2003dictionary, elad2006image, ramirez2010classification}. To learn a dictionary, one can apply any available solver such as MOD, KSVD and OLM. We choose to use a new dictionary learning method, which applies the BPG method proposed in \cite{xublock} to \eqref{spdict}. Compared to some state-of-the-art methods, the new algorithm is often faster and produces more faithful dictionaries. Though \eqref{spdict} is non-convex jointly with respect to $\mbfd$ and $\mbfy$, it is convex with respect to each of them while the other one is fixed. With this bi-convexity property, the BPG method is shown to generate a sequence globally converging to a stationary point of \eqref{spdict}.

\subsection{Block proximal gradient method}\label{sec:bpg}
Recently, \cite{xublock} characterized a class of \emph{multi-convex} problems
and proposed a BPG method
for solving these problems. For simplicity and our purpose, we review the method only for \emph{bi-convex} problems like \eqref{spdict}. Consider
\begin{equation}\label{eq:multi-convex}
\min_{\bfx,\bfy} f(\bfx,\bfy)+ r_x(\bfx)+r_y(\bfy),
\end{equation}
where $f$ is {differentiable} and convex with respect to either $\bfx$ or $\bfy$ by fixing the other one, and $r_x, r_y$ are  {extended-valued convex} functions.
At the $k$-th iteration of BPG, $\bfx$ and $\bfy$ are updated alternatively by
\begin{subequations}\label{eq:bpg}
\begin{align}
\bfx^k=&\argmin_{\bfx}\langle\nabla_\bfx f(\hat{\bfx}^{k},\bfy^{k-1}),\bfx-\hat{\bfx}^{k}\rangle+\frac{L_x^k}{2}\|\bfx-\hat{\bfx}^{k}\|_2^2+r_x(\bfx),\label{eq:bpg-x}\\
\bfy^k=&\argmin_{\bfy}\langle\nabla_\bfy f({\bfx}^{k},\hat{\bfy}^{k}),\bfy-\hat{\bfy}^{k}\rangle+\frac{L_y^k}{2}\|\bfy-\hat{\bfy}^{k}\|_2^2+r_y(\bfy),\label{eq:bpg-y}
\end{align}
\end{subequations}
where $L_x^k$ is a Lipschitz constant of $\nabla_x f(\bfx,\bfy^{k-1})$ with respect to $\bfx$, $\hat{\bfx}^{k}={\bfx}^{k-1}+\omega_x^{k}({\bfx}^{k-1}-{\bfx}^{k-2})$ denotes an extrapolated point with weight $\omega_x^{k}\ge0$, and $L_y^k$ and $\hat{\bfy}^k$ have the same meanings for $\bfy$. %It was demonstrated that a careful choice of $\omega_x^{k}$ and $\omega_y^k$ could significantly accelerate the algorithm; see Figure \ref{fig:accel}.

BPG is a variant of the block coordinate minimization (BCM) method (see \cite{Tseng-01} and the references therein), which updates $\bfx,\bfy$ cyclically by minimizing the objective with respect to one block of variables at a time while the other is fixed at its most recent value.
%\begin{equation}\label{eq:bcd}
%\bfx_i^k=\argmin_{\bfx_i}f(\bfx_{j<i}^k,\bfx_i,\bfx_{j>i}^{k-1})+r_i(\bfx_i), \text{ for }i=1,\cdots,s.
%\end{equation}
Though BCM decreases the objective faster, subproblems for BCM are usually much more difficult than those in \eqref{eq:bpg}. For simple $r_x$ and $r_y$, the updates in \eqref{eq:bpg} have closed form solutions.
%For instance, if $r_x(\bfx)=\delta_+(\bfx)$, an indicator function on nonnegative orthant which equals \emph{zero} if all components of $\bfx$ are nonnegative and $+\infty$ otherwise, then the update in \eqref{eq:bpg-x} can be explicitly written as
%$$\bfx^k=\max\left(0,\hat{\bfx}^k-\frac{1}{L_x^k}\nabla_\bfx f(\hat{\bfx}^{k},\bfy^{k-1})\right).$$

Under some boundedness assumptions, \cite{xublock} establishes subsequence convergence of the APG method. Further assuming the so-called Kurdyka-{\L}ojasiewicz (KL) property (see \cite{lojasiewicz1993geometrie, bolte2007lojasiewicz} for example), it shows that the sequence $\{(\bfx^k,\bfy^k)\}$ generated by \eqref{eq:bpg} globally converges to a stationary point of \eqref{eq:multi-convex} .

\subsection{Dictionary learning}
We learn a dictionary from training dataset $\mbfx$ via solving \eqref{spdict}. Let
$$\ell(\mbfd,\mbfy)=\frac{1}{2}\|\mbfd\mbfy-\mbfx\|_F^2$$ be the fidelity term in \eqref{spdict}. Applying \eqref{eq:bpg} to \eqref{spdict}, we alternatively update $\mbfd$ and $\mbfy$ by
\begin{subequations}\label{bpg-dl}
\begin{align}
\mbfd^k=&\argmin_{\mbfd\in\mcd}\langle\nabla_\mbfd\ell(\hat{\mbfd}^k,\mbfy^{k-1}),\mbfd-\hat{\mbfd}^k\rangle+\frac{L_d^k}{2}\|\mbfd-\hat{\mbfd}^k\|_F^2,\label{bpg-dl-d}\\
\mbfy^k=&\argmin_{\mbfy}\langle\nabla_\mbfy\ell({\mbfd}^k,\hat{\mbfy}^{k}),\mbfy-\hat{\mbfy}^k\rangle+\frac{L_y^k}{2}\|\mbfy-\hat{\mbfy}^k\|_F^2+\lambda\|\mbfy\|_1,\label{bpg-dl-y}
\end{align}
\end{subequations}
where $$\mcd=\{\mbfd:\|\bfd_i\|_2\le 1, i=1,\ldots,K\}$$
is the constraint set of $\mbfd$, $\hat{\mbfd}^k=\mbfd^{k-1}+\omega_d^k(\mbfd^{k-1}-\mbfd^{k-2})$ and $\hat{\mbfy}^k=\mbfy^{k-1}+\omega_y^k(\mbfy^{k-1}-\mbfy^{k-2})$ denote extrapolated points with $\omega_d^k,\omega_y^k\le 1$, and $L_d^k$ and $L_y^k$ are taken as Lipschitz constants of $\nabla_\mbfd\ell(\mbfd,\mbfy^{k-1})$ and $\nabla_\mbfy\ell(\mbfd^k,\mbfy)$ about $\mbfd$ and $\mbfy$ respectively.

The updates in \eqref{bpg-dl} can be explicitly written as
\begin{subequations}\label{update-dl}
\begin{align}
\mbfd^k=&\mcp_\mcd\left(\hat{\mbfd}^k-\frac{1}{L_d^k}\nabla_\mbfd\ell(\hat{\mbfd}^k,\mbfy^{k-1})\right),\label{update-dl-d}\\
\mbfy^k=&\mcs_{\lambda/L_y^k}\left(\hat{\mbfy}^k-\frac{1}{L_y^k}\nabla_\mbfy\ell({\mbfd}^k,\hat{\mbfy}^{k})\right),\label{update-dl-y}
\end{align}
\end{subequations}
where in \eqref{update-dl-d}, $\mcp_\mcd(\cdot)$ denotes the Euclidean projection to $\mcd$ defined for any $\mbfd$ as
$$\big(\mcp_\mcd(\mbfd)\big)_i=\frac{\bfd_i}{\max(1,\|\bfd_i\|_2)},~i=1,\ldots,K,$$
and in \eqref{update-dl-y}, $\mcs_\tau(\cdot)$ denotes soft-thresholding operator defined for any $\mbfy$ by
$$\big(\mcs_\tau(\mbfy)\big)_{ij}=\sign(y_{ij})\cdot\max(|y_{ij}|-\tau,0),~\forall~i,j.$$

Note that $\nabla_\mbfd\ell(\mbfd,\mbfy)=(\mbfd\mbfy-\mbfx)\mbfy^\top$ and
$$\|\nabla_\mbfd\ell(\mbfd,\mbfy)-\nabla_\mbfd\ell(\tilde{\mbfd},\mbfy)\|_F=\|(\mbfd-\tilde{\mbfd})\mbfy\mbfy^\top\|_F\le\|\mbfy\mbfy^\top\|\|\mbfd-\tilde{\mbfd}\|_F,~\forall ~\mbfd,\tilde{\mbfd},$$
where $\|\mbfa\|$ denotes matrix operator norm of $\mbfa$. Hence, $\|\mbfy\mbfy^\top\|$ is a Lipschitz constant of $\nabla_\mbfd\ell(\mbfd,\mbfy)$ about $\mbfd$. Throughout our numerical tests, we take
\begin{equation}\label{chooseL}
L_d^k=\|\mbfy^{k-1}(\mbfy^{k-1})^\top\|,\qquad L_y^k=\|(\mbfd^k)^\top\mbfd^k\|.\end{equation}
The extrapolation weights are taken as\footnote{In \eqref{omega}, the number ``0.9999'' can be replaced by any positive number less than \emph{one} to guarantee the global convergence of the algorithm. Numerically, the number that is closer to \emph{one} makes the algorithm converge faster. We observed that even if it was equal \emph{one}, the algorithm still converged.}
\begin{equation}\label{omega}
\omega_d^k=0.9999\min\left(\omega^k,\sqrt{\frac{L_d^{k-1}}{L_d^k}}\right),\qquad \omega_y^k=0.9999\min\left(\omega^k,\sqrt{\frac{L_y^{k-1}}{L_y^k}}\right),
\end{equation}
where $\omega^k=\frac{t_{k-1}-1}{t_k}$ with $t_0=1$ and $t_k=\frac{1}{2}\left(1+\sqrt{1+4t_{k-1}^2}\right)$. The weight $\omega^k$ has been used in FISTA \cite{BeckTeboulle2009}, showing that this kind of extrapolation significantly accelerates the proximal gradient method for convex composite problems. We observe that the extrapolation with weights in \eqref{omega} can also greatly speed up the BPG method for solving \eqref{spdict}.
%We compared our method with extrapolation (using weights as in \eqref{omega}) and without extrapolation (setting $\omega_x^k=\omega_y^k=0,~\forall k$) in Figure \ref{fig:accel}. The data was randomly generated in the same way as in section \ref{sec:syn}. It shows that the extrapolation technique greatly improve the convergence of BPG for \eqref{spdict}.

%\begin{figure}\caption{Comparison of BPG method with and without extrapolation on randomly generated data. ``with extrapolation'' uses the weight specified in \eqref{omega}, and ``no extrapolation'' sets $\omega_x^k=\omega_y^k=0,~\forall k$.}\label{fig:accel}
%\begin{center}
%\includegraphics[width=0.4\textwidth]{test_accel.pdf}
%\end{center}
%\end{figure}

To make the whole objective non-increasing, we redo the $k$-th iteration by setting $\omega_d^k=\omega_y^k=0$ (i.e., no extrapolation) if $F(\mbfd^k,\mbfy^k)>F(\mbfd^{k-1},\mbfy^{k-1})$,  where
$$F(\mbfd,\mbfy)=\frac{1}{2}\|\mbfd\mbfy-\mbfx\|_F^2+\lambda\|\mbfy\|_1$$
 is the objective of \eqref{spdict}. As shown in \cite{xublock}, the setting of $\omega_d^k=\omega_y^k=0$ guarantees $F(\mbfd^k,\mbfy^k)$ no greater than $F(\mbfd^{k-1},\mbfy^{k-1})$. The non-increasing property is not only required by global convergence, but also important to make the algorithm perform stably and converge rapidly. The pseudocode of our method is shown in Algorithm \ref{alg:bpg}.

\begin{algorithm}\caption{Block proximal gradient for dictionary learning}\label{alg:bpg}
{\small
\DontPrintSemicolon
\KwData{training samples $\mbfx$, parameter $\lambda>0$, and initial points $(\mbfd^{-1},\mbfy^{-1})=(\mbfd^{0},\mbfy^{0})$}
\For{$k=1,2,\cdots$}{
Set $L_d^{k}$ and $\omega_d^k$ by \eqref{chooseL} and \eqref{omega}, respectively.\;
Let $\hat{\mbfd}^k=\mbfd^{k-1}+\omega_d^k(\mbfd^{k-1}-\mbfd^{k-2})$ and get $\mbfd^k$ by \eqref{update-dl-d}.\;
Set $L_y^{k}$ and $\omega_y^k$ by \eqref{chooseL} and \eqref{omega}, respectively.\;
Let $\hat{\mbfy}^k=\mbfy^{k-1}+\omega_y^k(\mbfy^{k-1}-\mbfy^{k-2})$ and get $\mbfy^k$ by \eqref{update-dl-y}.\;
\If{$F(\mbfd^{k},\mbfy^k)>F(\mbfd^{k-1},\mbfy^{k-1})$}{
\lnlset{reupdate}{ReDo}Re-update $\mbfd^{k}$ and $\mbfy^k$ by \eqref{update-dl-d} and \eqref{update-dl-y} with $\hat{\mbfd}^{k}={\mbfd}^{k-1}$ and $\hat{\mbfy}^k=\mbfy^{k-1}$, respectively.\;
}
\If{Some stopping conditions are satisfied}{
Output $(\mbfd^k,\mbfy^k)$ and stop.
}
}
}
\end{algorithm}

\begin{remark}
Our algorithm uses proximal update for both $\mbfd$ and $\mbfy$. It differs from other methods such as KSVD and OLM which perform exact minimization to update $\mbfd$ and/or $\mbfy$. Maintaining closed form solutions for both $\mbfd$ and $\mbfy$-subproblems ensures the algorithm to have a lower per-iteration complexity, and the extrapolation technique lets it take a small number of iterations to achieve a faithful solution.
\end{remark}

\subsection{Convergence results}
Note that \eqref{spdict} is equivalent to
\begin{equation}\label{eqspdict}
\min_{\mbfd,\mbfy}\frac{1}{2}\|\mbfd\mbfy-\mbfx\|_F^2+\lambda\|\mbfy\|_1+\delta_\mcd(\mbfd),
\end{equation}
where $\delta_\mcd(\cdot)$ is the indicator function on $\mcd$. According to \cite{xublock}, the objective of \eqref{eqspdict} is semi-algebraic \cite{bierstone1988semianalytic} and has the KL property. In addition, the sequence $\{\mbfd^k\}$ is in the bounded set $\mcd$, and positive $\lambda$ makes $\{\mbfy^k\}$ bounded because otherwise the objective of \eqref{eqspdict} will blow up. Hence, $\{(\mbfd^k,\mbfy^k)\}$ has a finite limit point, and the Lipschitz constants specified in \eqref{chooseL} must be upper bounded. On the other hand, as long as $\{\mbfd^k\}$ and $\{\mbfy^k\}$ are uniformly away from origin, $L_d^k$ and $L_y^k$ are uniformly above \emph{zero}. Therefore, according to Theorem 2.8 of \cite{xublock}, we immediately have the following theorem.
\begin{theorem}\label{thm:convg}
Let $\{(\mbfd^k,\mbfy^k)\}$ be the sequence generated by Algorithm \ref{alg:bpg}. If both $\{\mbfd^k\}$ and $\{\mbfy^k\}$ are uniformly away from origin, then $(\mbfd^k,\mbfy^k)$ converges to a stationary point of \eqref{eqspdict} or equivalently \eqref{spdict}.
\end{theorem}

\section{Numerical results}\label{sec:numerical}
In this section, we first test Algorithm \ref{alg:bpg} for dictionary learning and compare it with KSVD \cite{aharon2006rm} and OLM \cite{mairal2009online} on synthetic data. Then we do a set of image recovery tests to show the effectiveness of model \eqref{un-model} and the adaptive method discussed in section \ref{sec:adp}.

\subsection{Synthetic test for dictionary recovery}\label{sec:syn}
This test compares Algorithm \ref{alg:bpg} with methods KSVD and OLM for dictionary learning. We chose KSVD and OLM because they appear to be most popular in the literature and their codes are both available online. In addition, they have been demonstrated efficient for many image processing tasks. There are other dictionary learning algorithms such as MOD \cite{engan2000multi} and recursive least squares \cite{skretting2010recursive}. However, we do not intend to exhaust all of them.

Following \cite{aharon2006rm}, we generated the test data as follows. We first generated a dictionary $\mbfd\in\mbr^{n\times K}$ with Matlab command \verb|randn(n,K)| and normalized each column of $\mbfd$ to have unit $\ell_2$-norm. Then we generated $p$ training samples in the $n$-dimensional space. Each sample is a linear combination of uniformly randomly selected $r$ columns of $\mbfd$, and the coefficients were Gaussian randomly generated. On the same data, we ran KSVD for \eqref{ksvd}, and both Algorithm \ref{alg:bpg} and OLM for \eqref{spdict}. In \eqref{ksvd} we set $s=r$, i.e., the true sparsity level was assumed, and in \eqref{spdict} we set $\lambda=0.5/\sqrt{n}$. Algorithm \ref{alg:bpg} was terminated as long as
$$\frac{|F(\mbfd^k,\mbfy^k)-F(\mbfd^{k+1},\mbfy^{k+1})|}{1+F(\mbfd^k,\mbfy^k)}\le 10^{-4}$$
was satisfied in three consecutive iterations or it ran over 1000 iterations. KSVD was run to 200 iterations, and OLM ran to the same time as that of Algorithm \ref{alg:bpg}. All other parameters for KSVD and OML were set to their default values.

We fixed $n=36$ and tested three different pairs of $(K,p)$. For each pair of $(K,p)$, sparsity level $r$ varied among $\{4,6,8,10,12\}$. The recovery of each atom $\bfd$ of the original dictionary $\mbfd$ was regarded successful if
$$\max_{1\le i\le K}\frac{|\bfd^\top\tilde{\bfd}_i|}{\|\bfd\|_2\|\tilde{\bfd}_i\|_2}\ge 0.99,$$
where $\tilde{\bfd}_i$ is the $i$-th column of an estimated dictionary $\tilde{\mbfd}$. The average running time and recovery rates of 50 independent runs are shown in Table \ref{table:syn}. From the table, we see that our method used much less time than KSVD with comparable recovery rates. When sparsity level $r$ is big (e.g., $r=12$) or the training samples are not so many (e.g., $p=20n$), our method got much higher recovery rates than those by KSVD. For the first two pairs of $(K,p)$, OLM tends to give lower rates than our method, and it may be because our method converges fast but OLM does not. However, in  the case $(K,p)=(4n,100n)$, we want to mention that OLM can give  results similar to ours if it is allowed to run  a very long time.

\begin{table}\caption{Average running time and recovery rates of 50 independent runs by Algorithm \ref{alg:bpg}, KSVD, and online learning method (OLM)}\label{table:syn}
\begin{center}
\resizebox{\textwidth}{!}{
\begin{tabular}{|c|cc|c|cc||cc|c|cc||cc|c|cc|}\hline
&\multicolumn{2}{|c|}{Algorithm \ref{alg:bpg}} & OLM & \multicolumn{2}{|c||}{KSVD}&\multicolumn{2}{|c|}{Algorithm \ref{alg:bpg}} & OLM & \multicolumn{2}{|c||}{KSVD}&\multicolumn{2}{|c|}{Algorithm \ref{alg:bpg}} & OLM & \multicolumn{2}{|c|}{KSVD}\\\hline
$r$ & time & rate(\%) & rate(\%) & time & rate(\%) & time & rate(\%) & rate(\%) & time & rate(\%) & time & rate(\%) & rate(\%) & time & rate(\%)\\\hline
&\multicolumn{5}{|c||}{$(K,p) = (2n,20n)$}&\multicolumn{5}{|c||}{$(K,p) = (2n,100n)$}&\multicolumn{5}{|c|}{$(K,p) = (4n,100n)$}\\\hline
 4 & 1.335 & 98.78  & 94.75 & 14.06 & 97.11 & 3.228 & 98.97 & 99.75 & 54.38 & 99.44 & 8.730 & 98.71 & 99.54 & 62.72 & 98.96
\\
 6 & 1.523 & 98.00  & 98.17 & 18.87 & 97.11 & 3.803 & 99.03 & 98.47 & 78.44 & 99.28 & 11.76 & 98.74 & 99.67 & 88.90 & 98.19
\\
 8 & 2.126 & 96.64  & 79.78 & 23.90 & 6.25 & 4.603 & 98.75 & 91.11 & 103.0 & 99.50& 15.45 & 98.42 & 99.44 & 116.2 & 98.21
\\
 10 & 2.975 & 94.22 & 52.22 & 29.06 & 0.00 & 5.919 & 98.11 & 76.00 & 128.6 & 97.25& 21.54 & 96.89 & 98.72 & 144.5 & 0.00
\\
 12 & 4.691 & 55.61 & 9.92 & 34.37 & 0.00 & 7.831 & 98.44  & 36.58 & 156.7 & 0.17& 30.18 & 83.29  & 70.39 & 174.4 & 0.00
\\\hline
\end{tabular}}
\end{center}
\end{table}

\subsection{Whole image recovery}\label{sec:imgrec}
This section tests the performance of Algorithms \ref{alg:rec} and \ref{alg:adp} on image recovery. Two different dictionaries were compared for Algorithm \ref{alg:rec}. One was an overcomplete DCT, generated in the same way as in \cite{aharon2006rm}. Another one was learned from 20,000 $8\times 8$ grayscale patches, that were 100 randomly extracted patches from each of the 200 images in the training set of the Berkeley segmentation dataset \cite{martin2001database}. For the learned dictionary, we first subtracted each training patch by its mean, and then trained a dictionary $\hat{\mbfd}$ using these zero-mean patches via solving \eqref{spdict} with $K=256$ by Algorithm \ref{alg:bpg}, where we chose $\lambda=0.8/\sqrt{n}$ to make the average nonzero number per column of $\mbfy$ about 8. Finally, we let $\mbfd=[\bfe,\hat{\mbfd}]\in\mbr^{64\times 257}$ and used $\mbfd$ in our tests, where $\bfe$ is a vector with all \emph{one}'s. Such an atom with constant components is called a DC in \cite{aharon2006rm}, which shows that the processed dictionary $\mbfd$ performs better than $\hat{\mbfd}$ for real-world image processing tasks. Here, we want to mention that for an image patch $\bfx$, if $\bfx-\text{mean}(\bfx)$ has a sparse representation under $\hat{\mbfd}$, i.e., $\bfx-\text{mean}(\bfx)=\hat{\mbfd}\bfy$ with sparse $\bfy$, then $\bfx=\text{mean}(\bfx)\bfe+\hat{\mbfd}\bfy$, which means $\bfx$ is sparse under $\mbfd$. Therefore, the above processing is reasonable. The used overcomplete DCT is also $64\times 257$, and its first column is a DC.
For Algorithm \ref{alg:adp}, we used the above $\mbfd$ as its initial dictionary, and updated the dictionary only once by learning a new one via Algorithm \ref{alg:bpg} using patches\footnote{Similarly, we subtracted every patch by its mean, and we augmented the learned dictionary by adding $\bfe$ as one more atom.} of the first-step estimated image, which is exactly the output of Algorithm \ref{alg:rec} using $\mbfd$. Then we used the updated dictionary to perform image recovery once more to get the final result.

\subsubsection*{Implementation}
In \eqref{un-model}, we took $\bfb=\mca(\mbfm)+\sigma\bm{\xi}$, where $\bm{\xi}\sim\mcn(0,\mbfi)$ is Gaussian noise, and $\sigma=\hat{\sigma}\|\mca(\mbfm)\|_2/\|\bm{\xi}\|_2$ throughout our tests, where $\hat{\sigma}$ varied among $\{0.01,0.05,0.10\}$. We took $\nu=\sigma$ for the first two kinds of $\mca$ and $\nu=0.1\sigma$ for the third kind of $\mca$. The definitions of different $\mca$'s are given in the next paragraph. In addition, we set all elements of $\bfw_{ij}$ to \emph{one} except its first component, which was set to \emph{zero}. Under this setting, using any DC as the first atom of $\mbfd$ would make no difference for the solution of \eqref{un-model}. Then, \eqref{un-model} was solved via YALL1 (version 1.4) \cite{zhang2010yall1}, for which we used Gaussian random starting point and $10^{-4}$ as its stopping tolerance. All other parameters of YALL1 were set to their default values. We chose YALL1 due to its high efficiency for solving \eqref{un-model} and easy call by providing operations of $\mca$ and $\mca^\top$.

Three different kinds of $\mca$ were tested. The first one did image inpainting and used the sampling operator $\mcp_\Omega$, which takes all pixels of its argument in $\Omega$ and zeros out all others. The adjoint of $\mcp_\Omega$ is to fill in the locations in $\Omega$ by its argument and other locations by zero. The second one did compressed image recovery and took $\mca$ as the composition of $\mcp_\Omega$ and two-dimensional complex-valued circulant operator $\mcc_2$, i.e., $\mca=\mcp_\Omega\circ\mcc_2$. We did the same normalization to $\mca$ as in \cite{XYO-circ2014}, which uses such $\mca$ for testing learned circulant sensing operators. Performing $\mcc_2$ on a matrix $\mbfm$ can be realized by one fast Fourier transform (FFT), one inverse FFT and some component-wise multiplications, and the adjoint of $\mcc_2$ is to do one fast Fourier transform (FFT), one inverse FFT and some component-wise divisions. The third kind of $\mca$ was a blurring operator with a $9\times 9$ kernel. We used two different kernels, which were generated by Matlab's commands \verb|fspecial('average',[9,9])| and \verb|fspecial('motion',10,45)| respectively. The implementation of a blurring operator can also be realized by one FFT, one inverse FFT, and some component-wise products. Hence, all the three kinds of $\mca$ can be easily realized in algorithms and in hardware. 

Our method processes a subset of nonoverlapping, covering patches together, and thus it recovers the whole image at a time. For $\mca=\mcp_\Omega$, one can denoise all overlapping patches independently since every measurement only involves one single pixel. However, this way would require noise information on each patch while our method only on the whole image. For the other two $\mca$'s, every measurement mixes more than one or even all image pixels, and one would not be able to process different patches independently. 

\subsubsection*{Results}
\begin{figure}\caption{Four tested images. From left to right: Castle, Lena, Plane, Boat}\label{fig:orig}
\begin{center}
\begin{tabular}{cccc}\\[-0.3cm]
\includegraphics[width=0.18\textwidth]{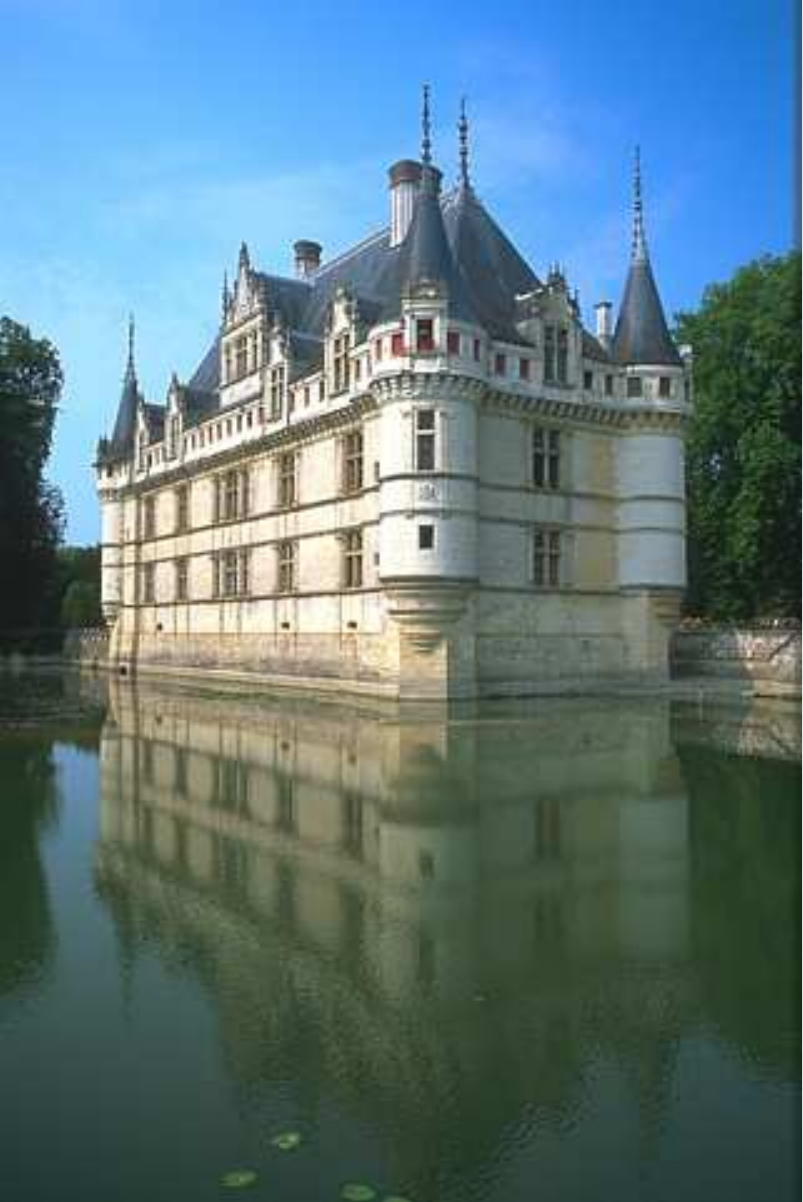} &
\includegraphics[width=0.22\textwidth]{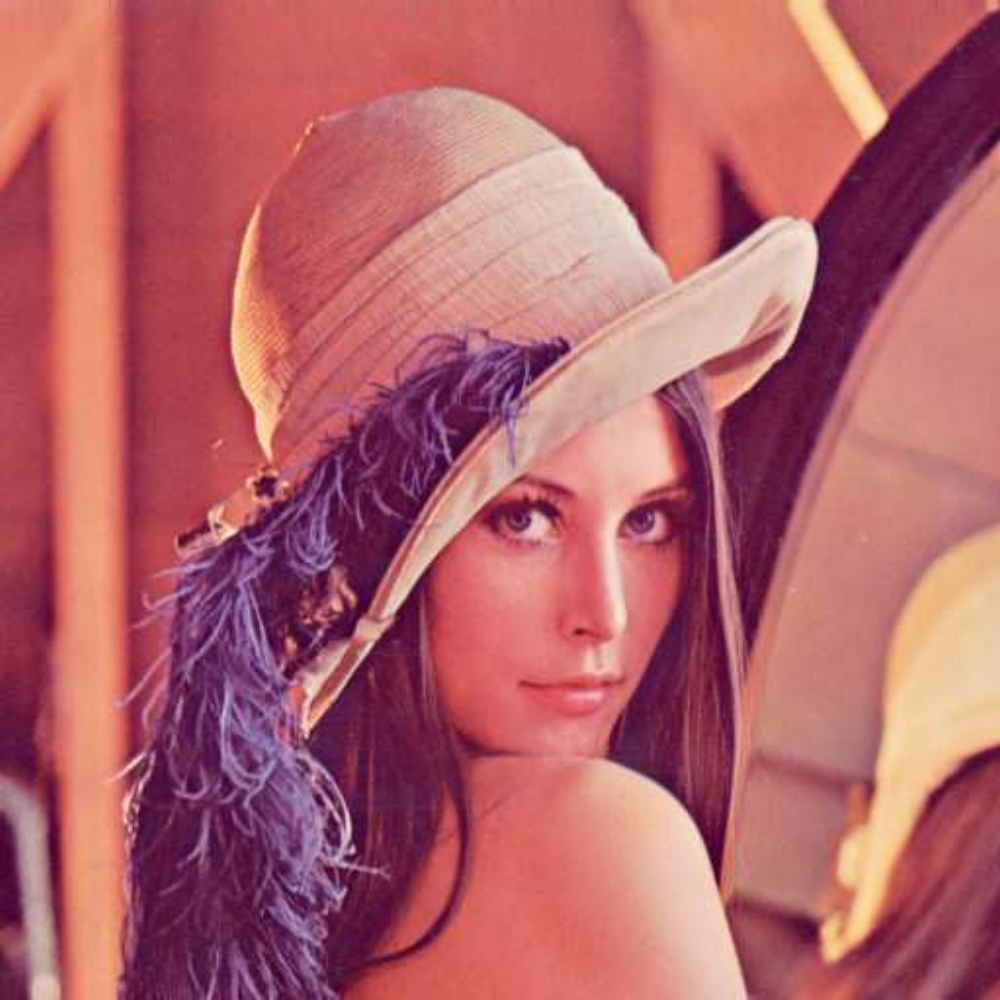} &
\includegraphics[width=0.27\textwidth]{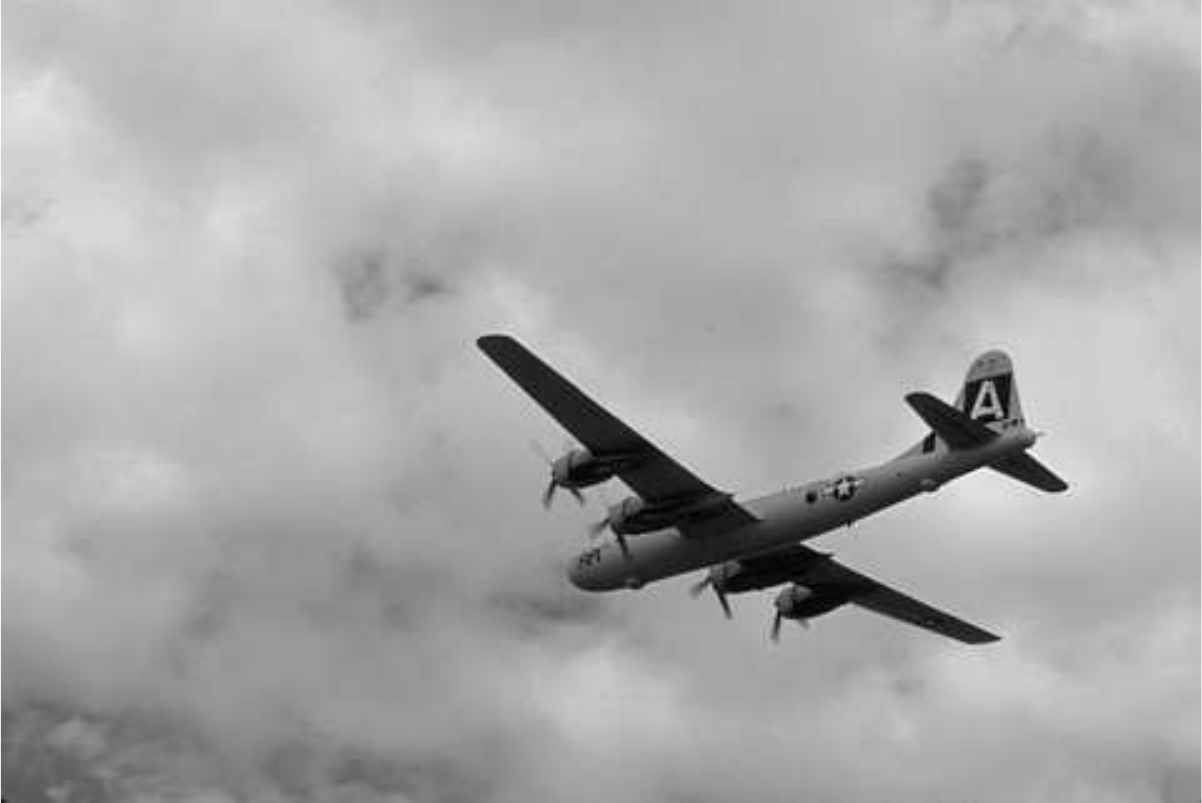} &
\includegraphics[width=0.22\textwidth]{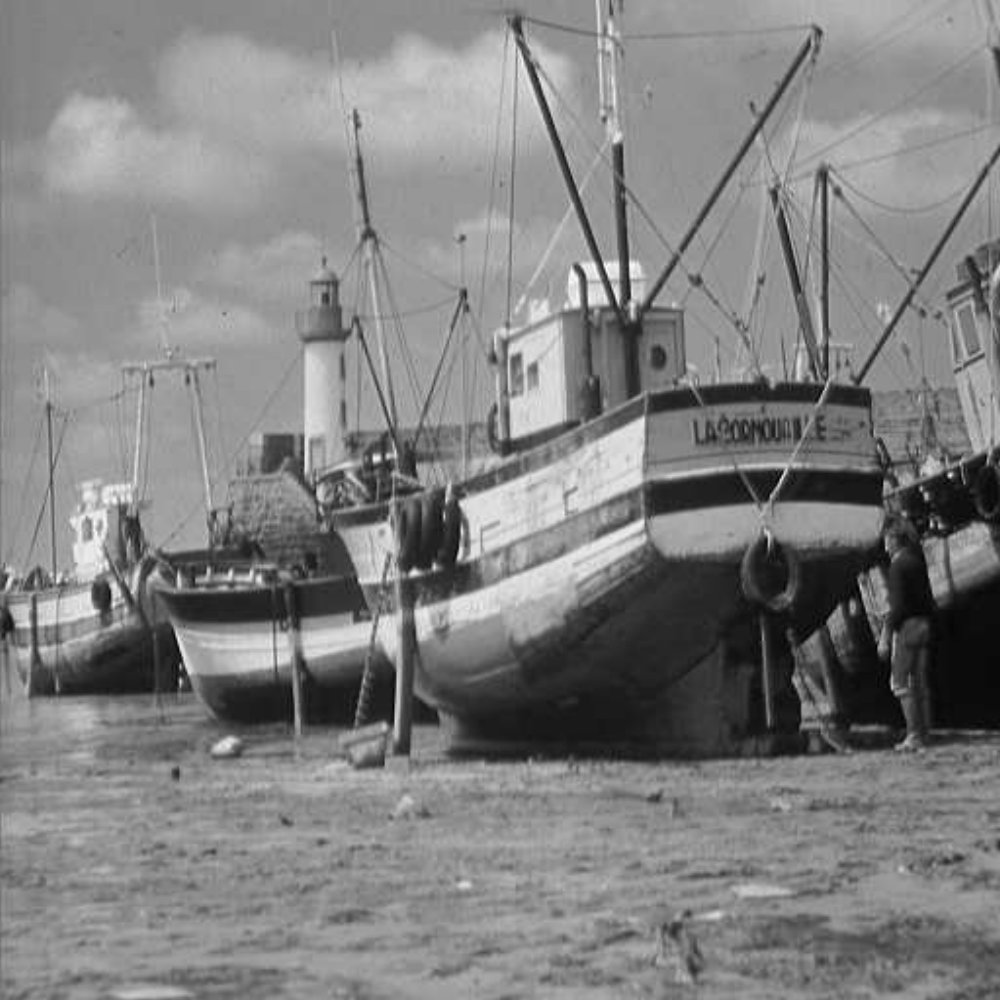}
\end{tabular}
\end{center}
\end{figure}

First, let us see how the averaging scheme in Algorithm \ref{alg:rec} improves the recovery performance. We tested it on the grayscale versions of Castle and Lena images shown in Figure \ref{fig:orig}, and both of the two images are unrelated to the training samples. We chose five different partitions, whose upper-left corner patches were $8\times 8$, $8\times 4$, $4\times 8$, $8\times 2$, and $2\times8$, respectively. (Recall that each partition is uniquely determined by its upper-left corner patch under Assumption \ref{assump:partition}.) For each partition, we solved \eqref{un-model} to obtain a recovered image. Let the recovered images be denoted by $\mbfm_1,\mbfm_2,\mbfm_3,\mbfm_4,\mbfm_5$. We compared PSNR values of the running average $\mbfm_j^{av}=\frac{1}{j}\sum_{i=1}^j\mbfm_i$ and the  $\mbfm_i$ that had the greatest PSNR among the five, denoted $\mbfm_{\text{best}}$. Table \ref{table:avg} lists the average results of five independent runs for four different $\mca$'s and noise level $\hat{\sigma}=1\%$. For the first two $\mca$'s, we took 30\% uniformly random pixels, i.e., $\text{SR}:=\frac{|\Omega|}{N_1N_2}=30\%$. From the results, we see that the averaging scheme consistently improves the recovery performance. Note that there are at most $n_1n_2$ different partitions under Assumption \ref{assump:partition}. We observed that the more different partitions we used, the better result could we  get by the averaging scheme. However, the rate of improvement drops as the number of partitions increases, as shown in Table \ref{table:avg}. For this reason, we only use three different partitions in the remaining experiments.

\begin{table}\caption{PSNR values of averaged images for $j=1,\ldots,5$. Every measurement vector contains 1\% Gaussian noise. For both image inpainting ($\mca=\mcp_\Omega$) and compressed imaging ($\mca=\mcp_\Omega\circ\mcc_2$), 30\% pixels were chosen uniformly at random.}\label{table:avg}
\begin{center}
\resizebox{\textwidth}{!}{\begin{tabular}{|c||cccccc||cccccc|}
\hline
Image & $\mbfm_{\text{best}}$ & $\mbfm_1^{av}$ & $\mbfm_2^{av}$ & $\mbfm_3^{av}$ & $\mbfm_4^{av}$ & $\mbfm_5^{av}$ & $\mbfm_{\text{best}}$ & $\mbfm_1^{av}$ & $\mbfm_2^{av}$ & $\mbfm_3^{av}$ & $\mbfm_4^{av}$ & $\mbfm_5^{av}$\\\hline
& \multicolumn{6}{|c||}{image inpainting} & \multicolumn{6}{|c|}{ compressed imaging}\\\hline
Castle & 25.23 & 25.05 &  25.80 &  26.21 &  26.36 &  26.48 & 34.13 & 34.04  & 34.86 &  35.27 &  35.44 &  35.57\\
Lena & 29.96 & 29.91  & 31.01 &  31.49 &  31.71 &  31.81 & 38.84 & 38.84  & 39.53 &  39.81 &  39.95 &  40.03\\\hline\hline
& \multicolumn{6}{|c||}{``average'' blurring} & \multicolumn{6}{|c|}{``motion'' blurring}\\\hline
Castle & 28.82 & 28.77 &  29.26 &  29.56 &  29.65 &  29.71 & 32.60 & 32.49  & 33.09 &  33.36 &  33.48 &  33.57\\
Lena & 32.26 & 32.22 &  32.79  & 33.09 &  33.20 &  33.29 & 36.55 & 36.55 &  37.17 &   37.44 &  37.56  & 37.63\\\hline\hline
\end{tabular}}
\end{center}
\end{table}

Next, we compare Algorithm \ref{alg:rec} with two different dictionaries and Algorithm \ref{alg:adp} on the four images shown in Figure \ref{fig:orig}. All of these images were unrelated to the learned dictionary $\mbfd$. To show the effectiveness of \eqref{un-model}, we also included a TV-based method for the first two $\mca$'s and an overlapping patch-based method for the third kind of $\mca$ in the comparison. The TV-based method solves
\begin{equation}\label{tv}\min_\mbfm\|\mbfm\|_\text{TV}+\frac{\gamma}{2}\|\mca(\mbfm)-\bfb\|_2^2,\end{equation}
where $\|\cdot\|_\text{TV}$ denotes TV semi-norm, and the overlapping patch-based method solves \eqref{asds}. We employed TVAL3 (version beta2.4) \cite{li2009tval3} to solve \eqref{tv}, and its default settings were used. The model \eqref{asds} was solved by the algorithm in \cite{dong2011image}, and its code was available online from the authors' webpage. We set its maximum number of iterations to $10^4$, which was sufficiently large to make the algorithm to solve \eqref{asds} to a high accuracy. In their code, the second group of local dictionaries were used, and we tuned the parameters \verb|par.tau| and \verb|par.c1| while all the other parameters were set to their default values. For color images, each of RGB channels was recovered independently.

For $\mca=\mcp_\Omega$, we tested $\text{SR}=30\%,~50\%$, and for $\mca=\mcp_\Omega\circ\mcc_2$, we tested $\text{SR}=10\%,~20\%,~ 30\%$. For each tested image, we chose three different partitions, whose upper-left corner patches were $8\times8$, $8\times4$, and $4\times8$, respectively. The same three partitions were used in both Algorithms \ref{alg:rec} and \ref{alg:adp}. Table \ref{table:mc} lists the average results of five independent trials by the compared methods for $\mca=\mcp_\Omega$, Table \ref{table:cs} for $\mca=\mcp_\Omega\circ\mcc_2$ and Table \ref{table:blur} for image deblurring. From the results, we see that Algorithm \ref{alg:rec} works better with learned $\mbfd$ than DCT except for the Castle image when $\mca$ is \verb|average| blurring operator and $\hat{\sigma}=10\%$. Our method with learned $\mbfd$ is consistently better for $\mca=\mcp_\Omega$ and much better for $\mca=\mcp_\Omega\circ\mcc_2$ than TV-based model \eqref{tv}. For both blurring operators, our method is better than that in \cite{dong2011image} for solving \eqref{asds} except when noise level $\hat{\sigma}=10\%$, the latter performs better on the Boat image for \verb|average| and the Castle image for both \verb|average| and \verb|motion|. In addition, Algorithm \ref{alg:adp} with adaptively updated dictionary makes improvement over Algorithm \ref{alg:rec} in all cases except for the Castle and Boat images when $\mca$ is \verb|average| blurring operator and $\hat{\sigma}=10\%$. The improvement usually increases as SR increases. It is reasonable since higher SRs give cleaner images, which further generate better dictionaries.

We provide open source codes on our websites and welcome the interested reader to try it on more datasets.

\begin{table}\caption{PSNR values of recovered images for image inpainting ($\mca=\mcp_\Omega$). From left to right, the results correspond to Algorithm \ref{alg:rec} with learned dictionary, Algorithm \ref{alg:rec} with DCT, Algorithm \ref{alg:adp}, and TV method, respectively. Bold is best.}\label{table:mc}
\begin{center}
{\footnotesize
\begin{tabular}{|c||cccc||cccc|}\hline
Image & \multicolumn{4}{|c||}{SR=30\%} & \multicolumn{4}{|c|}{SR=50\%}\\\hline
\multicolumn{9}{|c|}{noise level $\hat{\sigma}=1\%$}\\\hline\hline
Castle & 26.16 &  24.58 &  \textbf{26.37} &  25.05 & 29.30 &  27.41 &  \textbf{29.51} &  27.88\\
Lena &  31.40 &  28.57 &  \textbf{31.70} &  29.07 & 35.33 &  31.98 &   \textbf{35.44} &  32.43\\
Plane &  32.66  & 29.17  & \textbf{33.46} &  30.31 & 37.43 &  32.62 &  \textbf{38.56} &   33.53\\
Boat  & 28.49   & 25.79  & \textbf{29.14}  & 26.70 & 31.86  & 29.05 &  \textbf{32.48}  & 30.00\\\hline
\multicolumn{9}{|c|}{noise level $\hat{\sigma}=5\%$}\\\hline\hline
Castle & 26.09 & 24.57  &  \textbf{26.23} &  24.99 & 29.22 &  27.30 &  \textbf{29.47} &  27.68\\
Lena &  31.02 &  28.50 &  \textbf{31.22} &  28.91 & 34.49 &  31.64 &   \textbf{34.81} &  31.96\\
Plane &  32.29  & 29.18  & \textbf{32.54} &  30.11 & 36.84 &  32.58 &  \textbf{37.54} &   33.02\\
Boat  & 28.31   & 25.82  & \textbf{28.63}  & 26.62 & 31.67  & 28.88 &  \textbf{32.28}  & 29.70\\\hline
\multicolumn{9}{|c|}{noise level $\hat{\sigma}=10\%$}\\\hline\hline
Castle & 25.85 & 24.47  &  \textbf{25.98} &  24.81 & 28.76 &  27.02 &  \textbf{29.00} &  27.15\\
Lena &  30.34 &  28.25 &  \textbf{30.53} &  28.47 & 33.21 &  30.96 &   \textbf{33.41} &  30.77\\
Plane &  31.81  & 29.06  & \textbf{32.03} &  29.50 & 35.67 &  32.18 &  \textbf{36.27} &   31.57\\
Boat  & 27.78   & 25.62  & \textbf{28.07}  & 26.30 & 30.75  & 28.36 &  \textbf{31.30}  & 28.90\\\hline
\end{tabular}}
\end{center}
\end{table}

\begin{table}\caption{PSNR values of recovered images for compressed imaging  ($\mca=\mcp_\Omega\circ\mcc_2$). From left to right, the results correspond to Algorithm \ref{alg:rec} with learned dictionary, Algorithm \ref{alg:rec} with DCT, Algorithm \ref{alg:adp}, and TV method, respectively. Bold is best.}\label{table:cs}
\begin{center}
\resizebox{\textwidth}{!}{
\begin{tabular}{|c||cccc||cccc||cccc|}\hline
Image & \multicolumn{4}{|c||}{SR=10\%} & \multicolumn{4}{|c||}{SR=20\%} & \multicolumn{4}{|c|}{SR=30\%}\\\hline
\multicolumn{13}{|c|}{noise level $\hat{\sigma}=1\%$}\\\hline\hline
Castle &27.91 &  26.00  & \textbf{28.27}  & 23.35 & 32.00 &  30.73 &  \textbf{33.35} &  25.10 &  35.22 & 35.46  & \textbf{37.76}  & 27.27\\
Lena & 32.19 &  29.53 &  \textbf{32.36} &  25.75 &  36.41 &  33.86 &  \textbf{36.76} &  27.93 & 39.48 &  37.64 &  \textbf{40.06} &  28.79\\
Plane & 39.56  & 36.11  & \textbf{40.72}  & 30.45 &  42.76 &  40.88  & \textbf{44.15}  & 31.52 & 45.60 &  43.51 &  \textbf{46.37} &  32.70\\
Boat  & 28.80 &  26.08 &  \textbf{28.93}  & 24.67 & 32.48 &  30.00 &  \textbf{32.98}  & 27.21 &  34.64 &  33.58 &  \textbf{35.66} &  27.24\\\hline
\multicolumn{13}{|c|}{noise level $\hat{\sigma}=5\%$}\\\hline\hline
Castle & 27.59 & 25.78 & \textbf{27.87} & 23.20 & 31.19 & 29.77 & \textbf{31.97} & 24.74 & 33.38 & 32.77 & \textbf{34.58} & 26.71\\
Lena & 31.22 & 29.07 & \textbf{31.38} & 25.70 & 34.49 & 32.49 & \textbf{34.73} & 27.87 & 36.53 & 34.94 & \textbf{36.83} & 28.78\\
Plane & 36.92 & 32.59 & \textbf{37.52} & 29.92 & 39.28 & 37.08 & \textbf{40.01} & 31.76 & 40.45 & 38.87 & \textbf{41.17} & 31.83\\
Boat & 28.21 & 25.86 & \textbf{28.32} & 24.65 & 31.50 & 29.12 & \textbf{31.79} & 25.95 & 33.41 & 31.65 & \textbf{33.85} & 28.08\\\hline
\multicolumn{13}{|c|}{noise level $\hat{\sigma}=10\%$}\\\hline\hline
Castle & 26.79 & 25.23 & \textbf{26.97} & 22.82 & 30.00 & 28.66 & \textbf{30.50} & 25.03 & 31.72 & 30.86 & \textbf{32.44} & 25.31\\
Lena & 29.92 & 28.03 & \textbf{30.03} & 25.88 & 32.75 & 31.04 & \textbf{32.92} & 27.51 & 34.24 & 32.76 & \textbf{34.42} & 28.65\\
Plane & 34.33 & 30.14 & \textbf{34.73} & 28.37 & 36.69 & 33.84 & \textbf{37.15} & 30.24 & 37.41 & 35.70 & \textbf{37.84} & 31.19\\
Boat & 27.26 & 25.35 & \textbf{27.33} & 24.76 & 30.13 & 28.06 & \textbf{30.30} & 27.01 & 31.90 & 30 12 & \textbf{32.16} & 26.89\\\hline
\end{tabular}}
\end{center}
\end{table}

\begin{table}\caption{PSNR values of recovered images for image deblurring. From left to right, the results correspond to blurred image, Algorithm \ref{alg:rec} with learned dictionary, Algorithm \ref{alg:rec} with DCT, Algorithm \ref{alg:adp}, and the overlapping patch-based method in \emph{\cite{dong2011image}} for solving \eqref{asds}, respectively.}\label{table:blur}
\begin{center}
{\footnotesize
\begin{tabular}{|c||ccccc||ccccc|}\hline
Image & \multicolumn{5}{|c||}{``average'' blurring} & \multicolumn{5}{|c|}{``motion'' blurring}\\\hline
\multicolumn{11}{|c|}{noise level $\hat{\sigma}=1\%$}\\\hline\hline
Castle &22.37 &29.50 &  28.62 &  \textbf{29.56} &  28.69 & 23.24 &  33.28 &  33.06 &  \textbf{34.26} &  31.71\\
Lena & 26.00 & 33.01 &  32.05 &  \textbf{33.04} &  32.38  & 27.88 & 37.34 &  36.43 &  \textbf{37.58} &  35.14 \\
Plane & 27.88 & 37.27 &  34.60 &  \textbf{37.62} &   33.74 & 28.66 & 40.98 &  39.63 &  \textbf{41.55} &   35.35\\
Boat  & 23.36 & 31.45  & 30.14 &  \textbf{31.54}  & 30.70 & 24.64 & 34.79 &  34.11 &  \textbf{35.31} & 33.92\\\hline
\multicolumn{11}{|c|}{noise level $\hat{\sigma}=5\%$}\\\hline\hline
Castle & 22.03 & 26.39 &25.99 &\textbf{26.47} & 25.42 & 22.91 & 27.39 & 26.43 & \textbf{27.62} & 26.85\\
Lena & 25.85 & 29.41 & 29.02 & \textbf{29.60} & 28.86 & 27.63 & 30.37 & 29.10 & \textbf{31.16} & 30.89\\
Plane & 27.64 & 31.84 & 30.89 & \textbf{32.00} & 29.91 &28.36 &34.18 & 31.78 & \textbf{34.49} & 31.74\\
Boat & 23.27 & 27.86 & 27.17 & \textbf{27.95} & 27.18 & 24.51 & 28.93 & 27.62 & \textbf{29.17} & 28.67\\\hline
\multicolumn{11}{|c|}{noise level $\hat{\sigma}=10\%$}\\\hline\hline
Castle & 21.77 & 24.10 & 24.14 & 23.92 & \textbf{24.93} & 22.60 & 24.95 & 24.29 & 25.02 & \textbf{25.30}\\
Lena & 25.44 & 27.49 & 27.02 & \textbf{27.55} & 27.25 & 26.95 & 29.61 & 28.64 & \textbf{29.65} & 28.33\\
Plane & 26.95 & 30.24 & 29.03 & \textbf{30.39} & 28.13 & 27.54 & 32.50 & 30.57 & \textbf{32.83} & 29.01\\
Boat & 22.99 & 25.34 & 25.20 & 25.13 & \textbf{25.81} & 24.13 & 26.66 & 25.62 & \textbf{26.75} & 26.70\\\hline
\end{tabular}}
\end{center}
\end{table}

\section{Conclusions}\label{sec:conclusion}
Dictionary learning has been popularly applied to image denoising, super-resolution, classification and feature extraction. Various algorithms have been proposed for learning dictionaries to achieve different goals. In this paper, we focus on whole-image recovery and develop novel methods for learning dictionaries and then recovering images quickly and faithfully. Our algorithm not only has low per-iteration complexity and also converges fast. %On synthetic data, it was shown that our algorithm could recover more dictionary atoms than some state-of-the-art methods within the same training time. Furthermore, we used the trained dictionary to perform image recovery, which was achieved by solving a non-overlapping patch-based model.
In the algorithm, using non-overlapping patches and averaging across different subsets of  patches greatly reduce the variable freedom and are critical for fast and successful recovery. %In addition, we applied averaging scheme and adaptively updated the used dictionary to further improve recovery. Numerical results on real-world images demonstrated that the learned dictionary by our algorithm was superior over overcomplete DCT for natural image recovery. Our model performed not only much better than a TV-based model for image inpainting and compressive image recovery but also better than a state-of-the-art overlapping patch-based model for image deblurring.

%\bibliographystyle{plain}
%\bibliography{paperCol,Nonconvex,concentration,cs,inverse,my_publications}

\end{document}